\documentclass[preprint,review,12pt]{elsarticle}

\usepackage{amssymb}
\usepackage{amsmath}
\usepackage{mathrsfs}
\usepackage{dsfont}
\usepackage{booktabs}
\usepackage{multirow}
\usepackage{lineno}
\usepackage{hyperref}

\usepackage{xcolor}
\usepackage[onehalfspacing]{setspace}
\usepackage{algorithm}
\usepackage{algorithmic}
\usepackage{makecell}
\usepackage{subcaption}
\usepackage[normalem]{ulem}
\usepackage{cleveref}

\newcommand{\class}[1]{\mathcal{C}_{#1}}
\newcommand{\order}[1]{\mathcal{O}(#1)}
\newcommand{\classorder}[1]{\mathcal{O}(\mathcal{C}_{#1})}
\newcommand{\prob}[1]{\text{P}(#1)}

\newcommand{\nom}{Nominal}
\newcommand{\clm}{CLM}
\newcommand{\bet}{Beta}
\newcommand{\expo}{Exponential}
\newcommand{\triang}{Triangular}
\newcommand{\cdw}{CDWCE}
\newcommand{\slc}{SLACE}
\newcommand{\sord}{SORD}
\newcommand{\clmbeta}{CLM Beta}
\newcommand{\clmexp}{CLM Exponential}
\newcommand{\clmtriang}{CLM Triangular}
\newcommand{\clmcdwce}{CLM CDWCE}
\newcommand{\clmslc}{CLM SLACE}
\newcommand{\clmsord}{CLM SORD}
\newcommand{\diff}[1]{\text{d}#1}

\crefname{appendix}{}{}
\Crefname{appendix}{}{}

\journal{Computers and Electronics in Agriculture}

\graphicspath{{images/}}

\begin{document}
\begin{frontmatter}

\title{A novel ordinal multi-view aggregation scheme for oak defoliation}

\author[ccia]{Francisco Bérchez-Moreno}
\author[forestales]{Ricardo Enrique Hernández-Lambraño}
\author[ccia]{David Guijo-Rubio}
\author[alcala]{Víctor Manuel Vargas\corref{cor}}\cortext[cor]{Corresponding author. Departamento de Teoría de la Señal y Comunicaciones, Universidad de Alcala, Escuela Politécnica. Campus Universitario. Ctra. Madrid-Barcelona, Km. 33,600, Madrid, 28801.}  \ead{victor.vargas@uah.es}
\author[forestales,ersaf]{Francisco José Ruiz-Gómez}
\author[ccia]{Juan Carlos Fernández}
\author[forestales,ersaf]{Pablo González-Moreno}

\affiliation[ccia]{organization={Departamento de Ciencia de la Computación e Inteligencia Artificial, Universidad de Córdoba},
            addressline={Campus Universitario de Rabanales, Edificio Albert Einstein. Ctra. N-IV, Km. 396}, 
            city={Córdoba},
            postcode={14071}, 
            state={Andalucía},
            country={Spain}}

\affiliation[forestales]{organization={Department of Forest Engineering, Laboratory of Dendrochronology, Silviculture and Global Change – DendrodatLab, Universidad de Córdoba},
            addressline={Campus Universitario de Rabanales. Ctra. N-IV, Km. 396}, 
            city={Córdoba},
            postcode={14071}, 
            state={Andalucía},
            country={Spain}}

\affiliation[alcala]{organization={Departamento de Teoría de la Señal y Comunicaciones, Universidad de Alcala},
            addressline={Escuela Politécnica. Campus Universitario. Ctra. Madrid-Barcelona, Km. 33,600}, 
            city={Madrid},
            postcode={28805}, 
            state={Madrid},
            country={Spain}}

\affiliation[ersaf]{organization={ERSAF. Andalusian Institute for Earth System Research (IISTA), Universidad de Córdoba},
            addressline={Campus Universitario de Rabanales. Ctra. N-IV, Km. 396}, 
            city={Córdoba},
            postcode={14071}, 
            state={Andalucía},
            country={Spain}}

\begin{highlights}
\item Novel multi-view ensemble for ordinal defoliation estimation
\item Ordinal methods outperform nominal baselines in defoliation tasks
\item Three-view ensemble achieves best overall performance and robustness
\item Ground-level images capture complementary crown information
\item Framework improves scalable and consistent forest health monitoring
\end{highlights}

\begin{abstract}
Forest decline driven by climate and biotic stressors threatens ecosystem functioning, making accurate monitoring of tree health essential. In this work, we address tree defoliation estimation as an ordinal classification problem using ground-level imagery. We propose a novel multi-view ensemble framework that aggregates predictions from Convolutional Neural Networks (CNNs) trained on different perspectives of individual trees (north, south, and crown). This approach leverages complementary visual information while preserving modelling consistency through a homogeneous ensemble design.

A comprehensive evaluation is conducted by comparing multiple ordinal classification methods and analysing the contribution of each view and their combinations. Results show that modelling the ordinal structure of defoliation levels improves performance over nominal approaches, while the proposed multi-view ensemble consistently outperforms single-view and pairwise configurations. In particular, the three-view ensemble achieves the most robust and accurate predictions across all evaluation metrics.

These findings highlight the potential of combining Deep Learning (DL), Ordinal Classification (OC), and multi-view aggregation for scalable, consistent, and objective forest health assessment in complex ecosystems such as Mediterranean dehesas.
\end{abstract}


\end{frontmatter}

\section{Introduction}\label{sec:introduction}
Forests are essential regulators of the global climate system and a reservoir of terrestrial biodiversity, providing a wide range of ecosystem services including carbon sequestration, water regulation, soil protection, and socio-economic resources for millions of people  \cite{faoGlobalForestResources2025}. At the tree level, these ecosystem services critically depend on the integrity of canopy functioning, as tree crown represents the primary interface through which carbon assimilation, transpiration, and energy exchange with the atmosphere are regulated \cite{verheyenForestCanopiesNaturebased2024}. However, the accelerating pace of forest mortality observed worldwide poses a critical threat to these functions \cite{allenGlobalOverviewDrought2010, andereggClimatedrivenRisksClimate2020, carnicerWidespreadCrownCondition2011}. In forests, tree mortality is rarely an abrupt process, as it is very often linked to decline processes, with prolonged phases of physiological deterioration \cite{manion1992forest}, during which cumulative stress exposure progressively impairs canopy functioning and overall tree vitality \cite{sanguesa-barredaDropsNeedleProduction2023}. Complex interactions among climate stressors such as drought, rising temperatures, and biotic agents including insect outbreaks and pathogens have been related to widespread forest decline and increases in background and episodic mortality events \cite{acosta-munozUnravellingKeyFactors2025, hartmannClimateChangeRisks2022}. These mortality processes not only reshape forest dynamics \cite{mcdowellPervasiveShiftsForest2020} but also have cascading impacts on carbon cycling, landscape flammability, and regional hydrology, challenging the resilience of forest ecosystems in an era of rapid environmental change \cite{andereggClimatedrivenRisksClimate2020}.

Monitoring forest health detecting and quantifying decline symptoms are central to apply effective adaptive management strategies to protect forests from the effects of global change \cite{acosta-munozEvolutionParadigmShift2024}. Traditionally, tree condition assessment and canopy defoliation have relied on ground-based visual surveys using standardised protocols (e.g.,  crown condition surveys of the International Co-operative Programme on assessment and monitoring of air pollution effects on forests - ICP Forests, \cite{eichhornPartIVVisual2020}). These  approaches typically classify tree crowns into a limited number of ordinal defoliation classes, usually ranging from three to five levels, often  linked with specific agents \cite{eichhornPartIVVisual2020}. While such surveys provide detailed and ecologically meaningful information at the tree level,  they are time-consuming, require extensive expert training, and remain inherently subject to observer-related uncertainty, which substantially constrains their scalability and consistency across large spatial and temporal extents \cite{torresRoleRemoteSensing2021}.

Despite recent advances in aerial and satellite platforms, including satellital sensors , UAV platforms and LiDAR technology  \cite{ariza-salamancaIntegrationLandsatTimeSeries2019,lauschUnderstandingForestHealth2017,navarro-cerrilloIntegrationWorldView2Airborne2019}, the interpretation of tree condition from image data remains challenging due to complex canopy geometries, species variability, and costs of the imagery and sensor platforms. Airborne sensors capture only the aggregate reflectance of the top canopy layer, which may not reflect structural and physiological heterogeneity below the surface of the crown. Furthermore, in complex canopies, spectral signals are influenced by three-dimensional structure, understory reflectance, shadowing, and inter-species variability, leading to potential misrepresentation of tree health attributes if vertical variability is ignored. For example, relationships between spectral heterogeneity and vegetation complexity remain inconsistent across spatial scales and vegetation types, indicating that structural complexity can decouple remotely sensed metrics from actual ecological states \cite{canto-sansoresImportanceSpatialScale2024, saimunComprehensiveReviewTree2025}.
The proliferation of ground-level imaging combined with Artificial Intelligence (AI) offers a promising avenue to bridge this gap \cite{kalinDefoliationEstimationForest2019}, as terrestrial approaches can directly capture within-crown structural variation that is poorly resolved by above-canopy sensors. Ground-based photographs captured with smartphones or automatic field cameras provide rich textural and colour information on individual trees and their crowns. When coupled with DL models \cite{schmidhuber2015deep}, particularly CNNs \cite{khan2020survey}, these image datasets can be leveraged to classify health status, and estimate the crown defoliation of the tree \cite{beloiuIndividualTreeCrownDetection2023,kalinDefoliationEstimationForest2019}. CNNs trained on annotated tree images have demonstrated high accuracy in discriminating subtle foliage loss and stress symptoms that may be missed by coarser remote sensing data \cite{da2019estimating}, underscoring the potential of DL-driven ground-level analysis to complement airborne and satellite monitoring frameworks.

Meanwhile, CNNs exploit spatial patterns in image data through layers of convolution, pooling, and non-linear activations, allowing models to capture subtle variations in leaf density \cite{toda2019convolutional}. DL approaches extend this capability by stacking multiple layers and learning complex feature representations, often surpassing traditional machine learning methods in classification tasks. For example, CNN-based approaches have been successfully applied to crop defoliation assessment using UAV imagery, where models such as DefoNet demonstrate the effectiveness of DL over traditional machine learning methods in challenging field conditions \cite{zhang2022assessing}. Similarly, CNNs combined with multispectral UAV data have been used for forest monitoring tasks, including tree species classification and crown condition assessment in heterogeneous environments \cite{ecke2024towards}. Furthermore, ensemble learning techniques, in which predictions from multiple models are combined, can improve robustness and generalisation by mitigating model-specific biases, as demonstrated in recent works leveraging multi-model ensembles for multispectral data analysis in agricultural applications \cite{scutelnic2026multi}.

In this study, we propose a novel ordinal multi-view aggregation scheme, in which multiple CNNs trained on ground-level photographs from different orientations are aggregated to produce a single, coherent prediction for each tree. The proposed framework can be interpreted as a homogeneous multi-view ensemble, leveraging complementary information captured from different perspectives, improving the reliability of tree defoliation estimates and providing a scalable strategy for tree-level health assessment. In this context, tree defoliation assessment is formulated as an OC problem \cite{gutierrez2015ordinal}, as the target labels represent ordered levels of foliage loss rather than independent categories. Unlike standard nominal classification, ordinal approaches explicitly account for the inherent ordering between classes, enabling models to penalise misclassifications according to their severity. This is particularly relevant in defoliation estimation, where confusion between adjacent classes is less critical than large deviations. By incorporating this ordinal structure into the learning process, models can better capture the gradual nature of canopy degradation and produce more meaningful and consistent predictions.

Using as study case the iconic dehesa agro-silvopastoral systems of the south-west of the Iberian Peninsula, we aim to integrate ground-level imaging and novel OC approaches in a DL approach to estimate of crown defoliation levels. Dehesa ecosystems are of high soicioeconomical and ecological value but face persistent threats from the decline of holm oak (\textit{Quercus ilex }L. subsp. \textit{ballota} (Desf.) Samp.) and cork oak (\textit{Quercus suber} L.) \cite{morales-rodriguezChallengesMediterraneanFagaceae2025, desampaioepaivacamilo-alvesDeclineMediterraneanOak2013}. This decline is a multifatorial syndrome linked to a combination of abiotic factors (extreme climatic events and land-use changes), and cumulative biotic stressors, between them, the oomycete \textit{Phytophthora cinnamomi} Rands is considered the main factor triggering chronic disease and sudden death events. This combination of factors results in variable patterns of crown dieback and mortality across spatial and temporal scales \cite{sanchez-cuestaEnvironmentalDriversInfluencing2021}. Moreover, the heterogeneous structure of dehesa systems, which combine open tree canopy with pasture and shrub layers, presents unique challenges for conventional forest health assessment. All these factors give an advantage to classification techniques based on DL algorithms over the use of remote sensing for the unsupervised identification and classification of crown defoliation stages in holm and cork oaks in dehesas.  Specifically, the objectives of this study are:
\begin{itemize}
    \item Validate DL-based models for tree-level defoliation estimation from ground photographs.
    
    \item Compare a wide variety of ordinal classification approaches.
    
    \item Estimate the relevance of different orientations of the ground photographs.
    
    \item Enhance monitoring capacity for dehesa ecosystems through the integration of OC and ground imagery analysis.
    
    \item Support citizen science-based forest monitoring through automated tree crown status assessment, boosting knowledge of forest health conditions in the south-west of the Iberian Peninsula.
\end{itemize}

\section{Materials and methods}\label{sec:methodology}
This section describes the dataset collected for oak defoliation assessment, the DL methodologies considered, and the proposed ordinal multi-view aggregation scheme.

\subsection{Dataset}\label{subsec:dataset}

\begin{figure}
    \centering
    \includegraphics[width=0.75\linewidth]{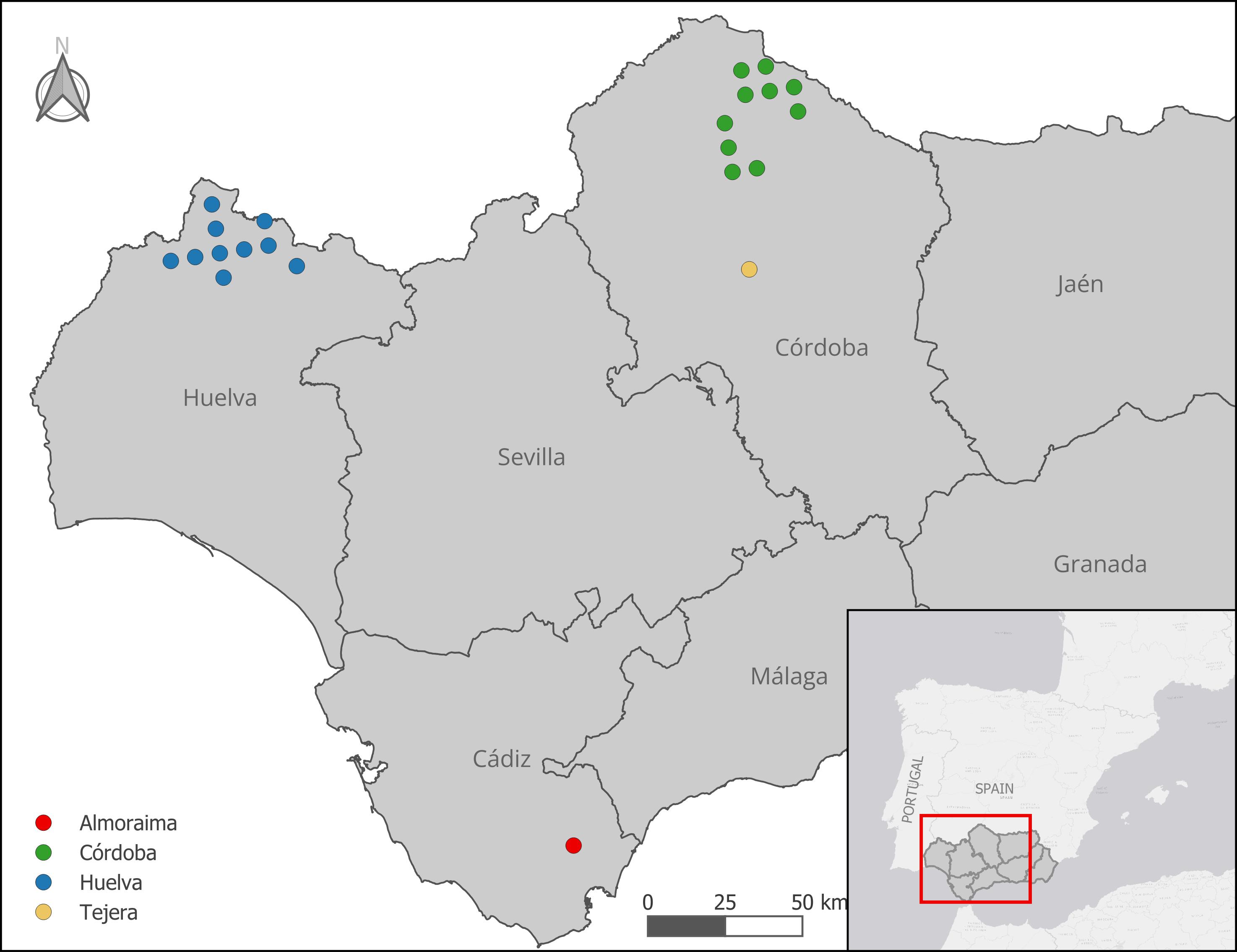}
    \caption{Map of the locations used for the assessment of crown defoliation of holm and cork oak. Blue points represent areas selected in Córdoba and Huelva for holm oak assessment, and in Cádiz for cork oak assessment. CRS: ETRS89 UTM /Zone 30 N}
    \label{fig:localizacion1}
\end{figure}

The study considered a total of 310 oak trees distributed across 22 sites in the Andalusian region (Southern Spain). Twenty of these sites were selected within the ``Red Andaluza de Seguimiento de Daños sobre Ecosistemas Forestales'' as part of the European network of monitoring plots of the International Co-operative Programme on Assessment and Monitoring of Air Pollution Effects on Forests (ICP Forest, Level I regional network). Specifically, we chose 10 sites in the western part of the region (Huelva) and other 10 in the east (Córdoba) to cover a wide environmental gradient (\Cref{fig:localizacion1}).  For each of these sites, 7 individual holm oak trees were chosen: the reference tree with best phytosanitary condition and six randomly selected to cover the range of defoliation levels in the plot (see further details in \cite{onoszkoDiversityPatternsHerbaceous2024}). This dataset comprised 136 trees gathered in spring 2021. Furthermore, we enriched this dataset with additional trees located in two  sites, one in Tejera in Córdoba (32 holm oak trees) and Almoraima in Cádiz (142 cork oak trees) where we established an intensive sampling protocol in Autumn 2024 (\Cref{fig:localizacion1}). 

\begin{table}
\centering
\caption{Defoliation classes for crown transparency percentages established in steps of 5 units and distribution of number of trees per health condition by expert visual assessment for each study area.}
\label{tab:dataset_thresholds}
\resizebox{\textwidth}{!}{
\begin{tabular}{cclcccc}
\toprule \toprule
Defoliation & Crown & \multirow{2}{*}{Description} & \multirow{2}{*}{Almoraima} & \multirow{2}{*}{Córdoba} & \multirow{2}{*}{Huelva} & \multirow{2}{*}{Tejera} \\
class & transparency (\%) & & & & &\\
\midrule
0 & 0\% – 10\% & Null defoliation & 18& 15& 5& 2\\
1 & 15\% – 25\% & Light defoliation & 46& 22& 22& 12\\
2 & 30\% – 60\% & Moderate defoliation & 42& 26& 28& 10\\
3 & 65\% – 95\% & Acute defoliation& 29& 6& 10& 2\\
4 & 100\% & Dead tree& 7& 2& 5& 6\\
\bottomrule \bottomrule
\end{tabular}
}

\end{table}

For each individual tree we recorded location, phytosanitary status and three different RGB images: one taken from the south, one from the north , and a third image taken from below, with the device positioned on the northern side of the crown at breast height and angled horizontally upwards. The defoliation of each tree was visually estimated using the methodology for crown assessment of the ICP Forest manual \cite{eichhornPartIVVisual2020}. Crown transparency was evaluated by two independent observers. A percentage value of crown transparency rounded to a multiple of 5 was set by each observer, and an average value agreed between the two observers was assigned to the tree. Then, each value was transformed into five ordinal classes following \cite{Ferreti2016} (see \Cref{tab:dataset_thresholds}). Note that the transparency intervals are closed, meaning that boundary values are included in the corresponding class. All information and imagery were recorded using an offline OpenDataKit \cite{Hartung2010ODK} form for android in a Tablet Samsung Galaxy Tab S9 (Samsung Electronics Co. Ltd, Seoul, South Korea). 

After the initial data collection, a preprocessing procedure was applied. From the original set of 310 trees, a total of 295 valid samples were retained, and all images were resized to $224 \times 224$ pixels to match the input requirements of the models.
Furthermore, due to the limited number of samples in the highest defoliation level (class 4 in \Cref{tab:dataset_thresholds}), the last two ordinal classes were merged into a single category. This results in a four-class ordinal classification problem, where the inherent ordering between classes is preserved.

Since each oak tree is represented by three distinct views (north, south, and crown), this allows the construction of three separate but aligned datasets. Since all views correspond to the same set of trees, they share an identical class distribution. The distribution of samples across training and test sets is summarised in \Cref{tab:dataset_distribution}. This distribution reflects a moderate degree of class imbalance, which is quantified using the Imbalance Ratio (IR). The IR is computed as:

\begin{equation}
    \label{eq:IR}
    \text{IR} = \frac{1}{Q} \sum^Q_{q=1} \frac{\sum_{i\neq q} N_i}{(Q-1) N_q},
\end{equation}

where $Q$ denotes the number of classes and $N_i$ represents the number of samples for the $i$-th class. This formulation yields a value of $1$ for perfectly balanced datasets, while values greater than $1$ indicate increasing imbalance. In our case, the dataset exhibits a moderate imbalance degree with an IR value of $1.277$.

\begin{table}[!ht]
\centering
\caption{Class distribution (number of patterns) for the training and test sets (identical across views), including the number of trees per plot.}
\label{tab:dataset_distribution}
\resizebox{\textwidth}{!}{
\begin{tabular}{ccccccccc}
\toprule\toprule
 & \multicolumn{2}{c}{Per view} \\
Class & Train & Test & Total & \% of patterns & Almoraima & Córdoba & Huelva & Tejera\\
\midrule
0 & 32 & 8  & 40  & 13.56\% & 18 & 15 & 5  & 2 \\
1 & 82 & 20 & 102 & 34.58\% & 46 & 22 & 22 & 12\\
2 & 85 & 21 & 106 & 35.93\% & 42 & 26 & 28 & 10\\
3 & 37 & 10 & 47  & 15.93\% & 29 & 6  & 10 & 2 \\
\bottomrule\bottomrule
\end{tabular}
}
\end{table}

\subsection{Modelling framework}\label{subsec:modeling}

The DL components of the neural networks investigated in this study are analysed, with particular attention to the output layer design and loss function, as these elements are crucial when modelling ordinal categories. Regarding the output layer, apart from the conventional nominal approach based on a softmax layer that models class probabilities independently, we consider an ordinal-aware threshold-based output derived from the Cumulative Link Model (CLM) \cite{vargas2020cumulative}, which enforces consistency by modelling cumulative class probabilities. 

Furthermore, the training process incorporates several ordinal-sensitive loss functions, namely Class Distance Weighted Cross Entropy (CDWCE) \cite{polat2022class}, Soft Ordinal Vectors (SORD) \cite{diaz2019soft}, and Soft Labels Accumulating Cross Entropy (SLACE) \cite{nachmani2025slace}. These loss functions introduce distance-aware penalisation mechanisms, assigning progressively larger penalties as predictions deviate further from the true class. Notably, SORD and SLACE inherently encode soft ordinal targets within their formulation, thereby integrating Soft Labelling (SL) directly into the optimisation process. Complementarily, we explore explicitly SL strategies based on triangular \cite{vargas2023soft}, beta \cite{vargas2022unimodal}, and exponential \cite{vargas2023exponential} probability distributions, which introduce controlled uncertainty around the true class while preserving the ordinal structure. These distributions are analysed both independently and in combination with the CLM output layer, enabling a comprehensive assessment of how uncertainty modelling and ordinal constraints interact during training in the context of oak defoliation.

All methodological variants are evaluated within a structured comparative framework that includes single-view models trained on each available tree perspective (north, south, and crown), pairwise combinations (north+south, north+crown, and south+crown), and a full three-view configuration. Within this task, we introduce a novel ordinal multi-view ensemble architecture composed of three identical classifiers, each trained on a distinct spatial view of the same tree instance. Diversity is induced through complementary spatial perspectives, as each classifier learns from a distinct view of the same tree instance, allowing the ensemble to capture view-specific variability while preserving modelling consistency. The final prediction is obtained through the combination of the outputs of the individual classifiers.

By embedding the different ordinal modelling strategies within this multi-view ensemble paradigm, the proposed framework enables an analysis of how output layer formulations, and loss functions (including SL schemes) interact across views. In this way, the multi (three)-view ensemble constitutes the central contribution of the study.

\subsubsection{Ordinal classification}\label{subsubsec:ordinal_classification}

Ordinal classification problems \cite{gutierrez2015ordinal} are those in which the objective is to classify patterns according to a categorical scale that reflects a natural order among the labels of the problem. Let $\mathcal{Y} = \{\class{1}, \class{2}, \ldots, \class{J}\}$ denote the set of $J$ ordered categories, arranged according to the relation $\class{1} \prec \class{2} \prec \cdots \prec \class{J}$, where $\prec$ represents the intrinsic ordering imposed by the application domain. 

Given an input space $\mathcal{X} \subseteq \mathds{R}^d$, with $d$ denoting the dimensionality of the feature space, each observation is described by a vector $\mathbf{x}_i \in \mathcal{X}$ and associated with a label $y_i \in \mathcal{Y}$. A supervised ordinal classification dataset is therefore defined as $\mathcal{D} = \langle \mathbf{X}, \mathbf{y} \rangle = \{(\mathbf{x}_i, y_i)\}_{i=1}^{N}$, where $\mathbf{X} = \langle \mathbf{x}_1, \ldots, \mathbf{x}_N \rangle$ represents the collection of input images, $\mathbf{y} = \langle y_1, \ldots, y_N \rangle$ the corresponding labels, i.e. the different levels of oak defoliation, and $N$ the total number of samples, i.e. oaks. The goal is to learn a predictive model that, given a new instance, produces an estimated label $\hat{y}_i \in \mathcal{Y}$.

In contrast to nominal classification, where categories are unordered, ordinal classification requires that prediction errors reflect the relative position of classes along the ordered scale. To this end, labels are mapped to their positions through a function $O(\cdot)$ such that each category $\class{j}$ is assigned the numerical value $\classorder{j} = j$, i.e., $O(y_i) = j$ whenever $y_i = \class{j}$ for $1 \leq j \leq J$. This representation enables the quantification of prediction discrepancies in a manner that preserves the ordinal semantics of the problem. Throughout this work, labels will be referred to either by their symbolic notation $\class{j}$ or by their numerical encoding $\classorder{j} = j$, depending on the context.

\subsubsection{Deep ordinal classification}\label{subsubsec:deep_ordinal_classification}
Deep ordinal classification extends standard DL frameworks to scenarios in which the output categories follow a natural ordering. In this context, two components play a central role in adapting deep neural networks to ordinal tasks: the output layer and the loss function. The output layer determines how class probabilities are parametrised and how the ordinal structure is encoded in the network outputs, while the loss function defines how prediction errors are penalised during training, often incorporating the ordinal relationships between categories into the optimisation process. Together, these elements guide the learning process so that the resulting model better reflects the ordered nature of the target variable.

Focusing on image-based OC, CNNs \cite{li2021survey} constitute a popular DL paradigm due to their ability to automatically extract hierarchical spatial features from raw pixel data. CNN architectures, such as AlexNet, VGG, and ResNet, have demonstrated strong performance across a wide range of visual recognition tasks, including OC problems, in which the output categories exhibit an inherent order. Their convolutional structure enables the modelling of local spatial dependencies, which are particularly relevant when subtle visual differences correspond to progressive ordinal levels.

In the specific case of oak defoliation detection, the problem involves analysing three complementary views (north, south, and crown) of the same tree, each capturing distinct spatial characteristics of foliage density and structural degradation. The defoliation levels follow a natural ordinal scale reflecting increasing severity, meaning that neighbouring categories share visual similarities while extreme categories exhibit more pronounced differences. CNNs are therefore particularly well suited to this problem, as they can capture fine-grained spatial variations within each view while learn feature representations that reflect gradual transitions between ordinal levels. This capability makes them an appropriate approach for modelling both the visual complexity and the ordered structure inherent to the oak defoliation classification problem. A more detailed description of the different ordinal output layer formulations, loss functions (including SL strategies) considered in this work can be found in \Cref{apx:a-deep-learning}.

\subsubsection{Ensemble learning}\label{subsubsec:ensemble_learning}
Ensemble methodologies aim to improve predictive performance by combining the outputs of multiple models. The fundamental idea is that combining diverse classifiers can yield more robust and better generalising predictions than relying on a single model \cite{opitz1999popular,hansen1990neural}. By reducing the impact of individual model biases and mitigating correlated errors, ensembles are particularly advantageous in complex classification scenarios. Ensembles can be broadly categorised into heterogeneous and homogeneous approaches. Heterogeneous ensembles combine predictions from different model architectures, exploiting complementary inductive biases. Homogeneous ensembles, in contrast, rely on multiple instances of the same base learner, introducing diversity through variations in data representation, input partitions, or training conditions. In multi-view problems such as defoliation detection, a homogeneous ensemble is particularly suitable, as using the same architectural framework across all views ensures modelling consistency, while diversity arises from the distinctive visual information contained in each perspective. This design isolates the contribution of each view without mixing it with architectural variability, facilitating clearer methodological comparisons and more controlled analysis.

Other ensemble variants increase diversity through feature transformations, data manipulations, or alternative problem decompositions. Regardless of the specific mechanism employed, the underlying objective remains the same: to promote complementary decision behaviours among base classifiers in order to enhance predictive stability. In the multi-view setting considered in this work, however, diversity naturally arises from the distinct visual perspectives available for each tree, which makes a homogeneous ensemble design particularly suitable for exploiting this complementary information while maintaining modelling consistency.

\subsection{Proposed ordinal multi-view aggregation scheme}\label{subsec:proposed_methodology}

\begin{figure}[!ht]
    \centering
    \includegraphics[width=0.83\textwidth]{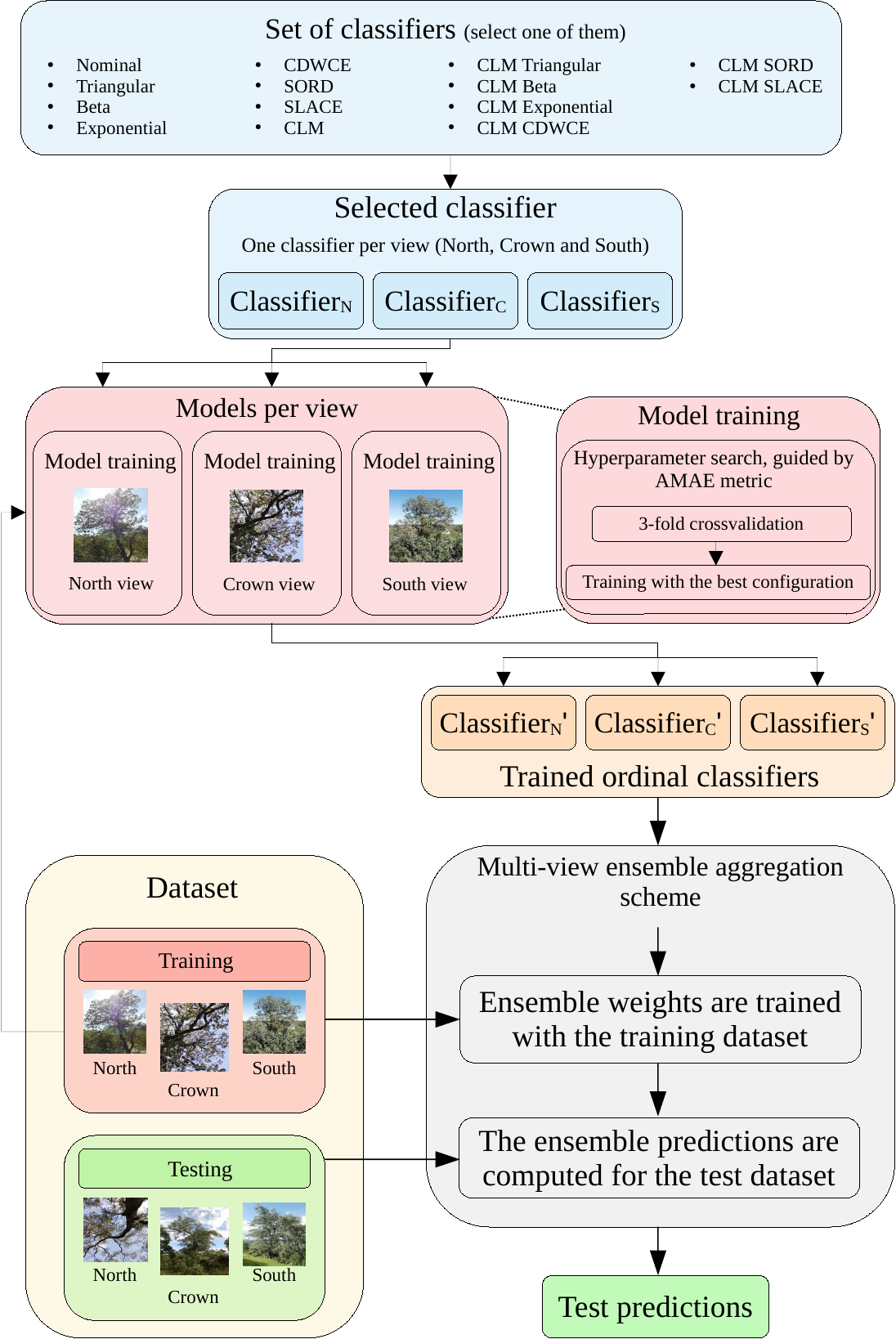}
    \caption{Flowchart of the proposed multi-view ensemble, illustrating the training of the classifiers, the adjustment of ensemble weights, and the final prediction phase.}
    \label{fig:ensemble}
\end{figure}

Our proposal presents a novel ordinal multi-view homogeneous ensemble where each classifier is trained with one of the views of the oaks. Each tree is observed from three distinct views, providing complementary visual information about foliage density. This procedure enable dedicated classifiers to specialise in the specific spatial patterns captured from each perspective. The aggregation of their outputs then integrates this complementary information at the decision level, improving robustness. As illustrated in \Cref{fig:ensemble}, the architecture can be instantiated either as a three-view ensemble, where all views (crown, north, and south) are combined, or as a set of pairwise ensembles, built from any of the three possible two-view combinations. This design provides flexibility in the way complementary visual information is integrated, enabling us to systematically evaluate whether performance gains are maximised when combining all three views or when focusing on specific pairwise combinations.

From the pool of methodologies considered in this study (see \Cref{fig:ensemble}), we select a single classifier, which is then independently trained on each view (north, crown, and south). This strategy is particularly important given that each view captures distinct visual characteristics, making it preferable to validate and optimise the models separately for each perspective. Once trained, these classifiers become the base learners of the proposed ensemble framework, which combines their outputs into a final class prediction. This combination process is performed in two stages over the three view-specific instances of the same base classifier. First, an optimisation mechanism determines the weight assigned to each classifier. This is achieved through a randomised search procedure, where candidate weight vectors are sampled from a uniform distribution and evaluated based on the ensemble's predictive performance. Second, the final prediction is derived by combining the individual outputs according to the learned weight vector $\mathbf{w}$. To formalise this process, we define the probability matrix $\mathbf{P}$, where each row corresponds to a classifier and each column to a class, so that $P_{ij}$ represents the probability assigned to class $j$ by classifier $i$. The final class probabilities are then obtained as the weighted sum of the rows of $\mathbf{P}$ according to the vector $\mathbf{w}$, and the predicted label is selected as the class with the highest aggregated probability.

\subsection{Experimental settings}\label{subsec:experiments}
This section outlines the procedures used for training and evaluating the different models for the proposed ordinal multi-view ensemble scheme, as well as the performance metrics employed at each stage. All experiments have been implemented using the \texttt{dlordinal} Python library \cite{berchez2025dlordinal}, which is openly accessible via GitHub\footnote{\url{https://github.com/ayrna/dlordinal}}.

\subsubsection{Compared methodologies and training setup}
For the experimental phase, and in order to determine the most suitable approach to be used within the proposed ensemble approach, a set of 14 methodologies are evaluated in this study. All methods are derived from a common baseline model built upon a ResNet18 architecture \cite{he2016deep}, trained using the Categorical Cross-Entropy (CCE) loss and a softmax output layer. From this nominal configuration, 13 additional variants are developed by modifying either the loss function, the output layer, or both, to incorporate ordinal information into the learning process. The full list of compared methodologies is summarised in \Cref{tab:compared_methodologies}.

\begin{table}[!ht]
    \centering
    \caption{Methodologies comparison. All of them include ResNet18 as their network architecture, varying the network loss and output layer to address ordinality.}
    \label{tab:compared_methodologies}
    \small
    \vspace{0.3cm}
    \begin{tabular}{lll}
        \toprule \toprule
        \textbf{Methodologies}     & \textbf{Loss function}                  & \textbf{Output Layer} \\
        \midrule
        \nom            & CCE                                 & Softmax      \\
        \triang         & CCE + SL (Triangular distribution)  & Softmax      \\
        \bet            & CCE + SL (Beta distribution)        & Softmax      \\
        \expo           & CCE + SL (Exponential distribution) & Softmax      \\
        \cdw            & CDWCE                               & Softmax      \\
        \sord           & SORD                                & Softmax      \\
        \slc            & SLACE                               & Softmax      \\
        \clm            & CCE                                 & CLM          \\
        \clmtriang      & CCE + SL (Triangular distribution)  & CLM          \\
        \clmbeta        & CCE + SL (Beta distribution)        & CLM          \\
        \clmexp         & CCE + SL (Exponential distribution) & CLM          \\
        \clmcdwce       & CDWCE                               & CLM          \\
        \clmsord        & SORD                                & CLM          \\
        \clmslc         & SLACE                               & CLM          \\
        \bottomrule \bottomrule
    \end{tabular}
\end{table}

The training process consists of fitting each model configuration to predict oak defoliation using 20 distinct random seeds to account for variability and ensure robustness. An initial stratified holdout split (80\% for training, and 20\% for test) was created and kept constant across all experiments to serve as the base partition. For each seed, a stratified resampling of the training subset was applied, maintaining the original class distribution to enhance generalisation. 

Hyperparameter tuning for each methodology was performed using a cross-validation procedure, guided by the Average Mean Absolute Error (AMAE) metric \cite{baccianella2009evaluation}. AMAE evaluates the mean absolute error independently for each class and then averages the results, making it particularly suitable for ordinal problems with imbalanced class distributions. A detailed description of the cross-validation setup and parameter ranges is provided in \Cref{apx:crossvalidation}.

\subsubsection{Models' evaluation}
To specifically evaluate the performance of the models capturing the ordinal structure of the task, we focus on two ordinal-aware metrics: the AMAE introduced before, and the Quadratic Weighted Kappa (QWK). In addition,  accuracy as an standard nominal metrics is also reported to provide a complementary perspective. A detailed description of all these metrics and their formal definitions can be found in \Cref{apx:metrics}. All performance metrics are estimated across the 14 compared methods, combination of image views and the 4 sites. Furthermore,  to assess whether the differences in the three metrics are statistically significant across the considered factors, we conduct a two-way ANOVA (ANOVA II) with Method and View as the independent variables (\ref{apx:statistical_analysis}). Finally a Tukey HSD test is used to statistically compare the 14 methodologies.

\section{Results}\label{sec:results}
This section presents the experimental results of the proposed approach. First, the performance of the 14 compared methodologies is analysed, assessing the impact of incorporating ordinal information. Then, the influence of combining multiple image views (crown, north, and south) within the multi-view ensemble framework is examined. Finally, results are reported by study site.

\subsection{Performance across models}\label{subsec:resultsmodels}

\begin{table}[htpb]
    \caption{Test results for the QWK, AMAE, and accuracy metrics, presented for each method and each individual view or group of views. Experiments involving a single-view use a single method, while those involving multiple views employ the ordinal multi-view ensemble of methods, one per view. The best result for each view is shown in bold, and the second-best is shown in italics. The method achieving the best overall value for each performance metric is double underscored, while the second best is single underscore. Metrics marked with $\uparrow$ are to be maximised and those with $\downarrow$ are to be minimised.}
    \centering
    \tiny
    \label{tab:results}
    \resizebox{\linewidth}{!}{
    \begin{tabular}{l*{8}{c@{\hskip 0.08cm}}}
    \toprule\toprule
     & Crown & North & South & \makecell{Crown+\\North} & \makecell{Crown+\\South} & \makecell{North+\\South} & \makecell{Crown+\\North+\\South} \\
    \midrule
    \multicolumn{8}{c}{QWK ($\uparrow$)}\\
    \midrule
    \nom & $\mathit{0.452_{0.18}}$ & $0.425_{0.23}$ & $0.409_{0.16}$ & $\mathbf{0.555_{0.10}}$ & $0.533_{0.12}$ & $0.519_{0.16}$ & $0.586_{0.08}$ \\
    \triang & $0.365_{0.15}$ & $0.270_{0.20}$ & $0.341_{0.23}$ & $0.464_{0.14}$ & $0.504_{0.13}$ & $0.432_{0.19}$ & $0.549_{0.11}$ \\
    \bet & $0.358_{0.19}$ & $0.368_{0.21}$ & $0.393_{0.14}$ & $0.497_{0.16}$ & $0.481_{0.14}$ & $0.523_{0.11}$ & $0.559_{0.11}$ \\
    \expo & $0.408_{0.16}$ & $0.386_{0.18}$ & $0.357_{0.17}$ & $0.530_{0.12}$ & $0.520_{0.12}$ & $0.507_{0.13}$ & $0.576_{0.10}$ \\
    \cdw & $\mathbf{0.454_{0.16}}$ & $0.377_{0.21}$ & $0.404_{0.23}$ & $0.546_{0.13}$ & $\mathbf{0.573_{0.10}}$ & $0.502_{0.16}$ & $\mathit{\uline{0.610_{0.09}}}$ \\
    \sord & $0.431_{0.13}$ & $0.404_{0.19}$ & $\mathit{0.435_{0.19}}$ & $0.536_{0.13}$ & $0.530_{0.11}$ & $0.523_{0.15}$ & $0.589_{0.10}$ \\
    \slc & $0.331_{0.19}$ & $0.342_{0.20}$ & $0.286_{0.20}$ & $0.478_{0.16}$ & $0.450_{0.17}$ & $0.458_{0.15}$ & $0.524_{0.15}$ \\
    \clm & $0.408_{0.15}$ & $0.476_{0.15}$ & $\mathbf{0.483_{0.18}}$ & $\mathit{0.553_{0.11}}$ & $\mathit{0.546_{0.16}}$ & $\mathbf{0.568_{0.13}}$ & $\mathbf{\uuline{0.613_{0.10}}}$ \\
    \clmtriang & $0.401_{0.17}$ & $\mathbf{0.483_{0.13}}$ & $0.409_{0.22}$ & $0.511_{0.13}$ & $0.513_{0.14}$ & $0.537_{0.11}$ & $0.546_{0.12}$  \\
    \clmbeta & $0.414_{0.13}$ & $\mathit{0.481_{0.10}}$ & $0.352_{0.19}$ & $0.506_{0.09}$ & $0.481_{0.15}$ & $0.509_{0.09}$ & $0.531_{0.10}$ \\
    \clmexp & $0.403_{0.14}$ & $0.349_{0.20}$ & $0.430_{0.19}$ & $0.486_{0.13}$ & $0.523_{0.14}$ & $\mathit{0.541_{0.13}}$ & $0.574_{0.11}$  \\
    \clmcdwce & $0.311_{0.16}$ & $0.377_{0.21}$ & $0.380_{0.21}$ & $0.463_{0.12}$ & $0.452_{0.15}$ & $0.483_{0.11}$ & $0.493_{0.10}$  \\
    \clmsord & $0.405_{0.14}$ & $0.430_{0.17}$ & $0.370_{0.20}$ & $0.495_{0.13}$ & $0.483_{0.14}$ & $0.513_{0.15}$ & $0.552_{0.12}$  \\
    \clmslc & $0.354_{0.18}$ & $0.396_{0.18}$ & $0.420_{0.22}$ & $0.485_{0.14}$ & $0.502_{0.15}$ & $0.478_{0.18}$ & $0.522_{0.13}$  \\
    \midrule
    \multicolumn{8}{c}{AMAE ($\downarrow$)}\\
    \midrule
    \nom & $\mathbf{0.678_{0.19}}$ & $0.742_{0.33}$ & $\mathit{0.731_{0.16}}$ & $\mathbf{0.557_{0.11}}$ & $\mathit{0.595_{0.15}}$ & $\mathit{0.598_{0.16}}$ & $0.527_{0.10}$  \\
    \triang & $0.812_{0.18}$ & $0.965_{0.32}$ & $0.862_{0.32}$ & $0.685_{0.18}$ & $0.644_{0.17}$ & $0.723_{0.23}$ & $0.588_{0.15}$ \\
    \bet & $0.852_{0.28}$ & $0.856_{0.31}$ & $0.801_{0.21}$ & $0.659_{0.21}$ & $0.683_{0.21}$ & $0.624_{0.14}$ & $0.576_{0.13}$  \\
    \expo & $0.785_{0.21}$ & $0.797_{0.22}$ & $0.870_{0.26}$ & $0.620_{0.15}$ & $0.631_{0.16}$ & $0.655_{0.16}$ & $0.562_{0.13}$  \\
    \cdw & $\mathit{0.691_{0.20}}$ & $0.837_{0.32}$ & $0.813_{0.35}$ & $\mathit{0.577_{0.17}}$ & $\mathbf{0.555_{0.14}}$ & $0.651_{0.20}$ & $\mathit{\uline{0.514_{0.12}}}$ \\
    \sord & $0.722_{0.17}$ & $0.785_{0.28}$ & $0.746_{0.24}$ & $0.592_{0.14}$ & $0.601_{0.11}$ & $0.633_{0.20}$ & $0.544_{0.11}$  \\
    \slc & $0.820_{0.20}$ & $0.873_{0.33}$ & $0.918_{0.29}$ & $0.660_{0.19}$ & $0.691_{0.21}$ & $0.695_{0.23}$ & $0.595_{0.19}$  \\
    \clm & $0.769_{0.18}$ & $\mathbf{0.693_{0.20}}$ & $\mathbf{0.703_{0.23}}$ & $0.592_{0.16}$ & $0.599_{0.19}$ & $\mathbf{0.583_{0.17}}$ & $\mathbf{\uuline{0.513_{0.11}}}$ \\
    \clmtriang & $0.804_{0.24}$ & $\mathit{0.706_{0.18}}$ & $0.824_{0.32}$ & $0.673_{0.18}$ & $0.650_{0.17}$ & $0.646_{0.17}$ & $0.622_{0.15}$ \\
    \clmbeta & $0.788_{0.19}$ & $0.713_{0.13}$ & $0.878_{0.31}$ & $0.669_{0.13}$ & $0.689_{0.22}$ & $0.676_{0.13}$ & $0.631_{0.15}$ \\
    \clmexp & $0.783_{0.18}$ & $0.874_{0.27}$ & $0.786_{0.27}$ & $0.683_{0.17}$ & $0.648_{0.19}$ & $0.634_{0.19}$ & $0.578_{0.15}$ \\
    \clmcdwce & $0.930_{0.23}$ & $0.864_{0.30}$ & $0.853_{0.32}$ & $0.721_{0.14}$ & $0.731_{0.20}$ & $0.694_{0.16}$ & $0.676_{0.12}$  \\
    \clmsord & $0.778_{0.19}$ & $0.768_{0.27}$ & $0.859_{0.30}$ & $0.658_{0.17}$ & $0.680_{0.19}$ & $0.645_{0.22}$ & $0.575_{0.13}$  \\
    \clmslc & $0.865_{0.28}$ & $0.830_{0.27}$ & $0.795_{0.34}$ & $0.682_{0.20}$ & $0.648_{0.21}$ & $0.700_{0.28}$ & $0.617_{0.19}$  \\
    \midrule
    \multicolumn{8}{c}{Accuracy ($\uparrow$)}\\
    \midrule
    \nom & $\mathbf{0.496_{0.11}}$ & $\mathbf{0.491_{0.13}}$ & $0.456_{0.10}$ & $\mathbf{0.564_{0.07}}$ & $\mathit{0.532_{0.09}}$ & $\mathit{0.537_{0.09}}$ & $\mathit{\uline{0.577_{0.07}}}$ \\
    \triang & $0.437_{0.08}$ & $0.395_{0.12}$ & $0.411_{0.14}$ & $0.501_{0.09}$ & $0.519_{0.10}$ & $0.467_{0.13}$ & $0.547_{0.09}$ \\
    \bet & $0.438_{0.12}$ & $0.430_{0.12}$ & $0.435_{0.09}$ & $0.519_{0.11}$ & $0.501_{0.10}$ & $0.525_{0.10}$ & $0.545_{0.09}$ \\
    \expo & $0.460_{0.11}$ & $0.463_{0.09}$ & $0.426_{0.10}$ & $0.529_{0.09}$ & $0.524_{0.10}$ & $0.512_{0.08}$ & $0.558_{0.08}$ \\
    \cdw & $\mathit{0.487_{0.12}}$ & $0.458_{0.13}$ & $0.450_{0.14}$ & $\mathit{0.555_{0.10}}$ & $\mathbf{0.550_{0.10}}$ & $0.516_{0.09}$ & $\mathbf{\uuline{0.579_{0.08}}}$ \\
    \sord & $0.474_{0.10}$ & $0.448_{0.12}$ & $0.442_{0.11}$ & $0.533_{0.10}$ & $0.514_{0.09}$ & $0.506_{0.09}$ & $0.542_{0.08}$ \\
    \slc & $0.425_{0.11}$ & $0.421_{0.13}$ & $0.414_{0.12}$ & $0.498_{0.12}$ & $0.493_{0.11}$ & $0.498_{0.11}$ & $0.542_{0.12}$ \\
    \clm & $0.426_{0.09}$ & $\mathit{0.489_{0.09}}$ & $\mathbf{0.485_{0.10}}$ & $0.536_{0.08}$ & $0.517_{0.10}$ & $\mathbf{0.540_{0.09}}$ & $0.570_{0.07}$ \\
    \clmtriang & $0.421_{0.10}$ & $0.478_{0.08}$ & $0.434_{0.11}$ & $0.485_{0.09}$ & $0.496_{0.09}$ & $0.502_{0.08}$ & $0.508_{0.08}$ \\
    \clmbeta & $0.424_{0.08}$ & $0.461_{0.07}$ & $0.406_{0.11}$ & $0.485_{0.06}$ & $0.487_{0.10}$ & $0.481_{0.07}$ & $0.511_{0.07}$ \\
    \clmexp & $0.423_{0.09}$ & $0.401_{0.10}$ & $0.432_{0.11}$ & $0.486_{0.08}$ & $0.492_{0.10}$ & $0.501_{0.08}$ & $0.537_{0.08}$ \\
    \clmcdwce & $0.396_{0.07}$ & $0.436_{0.11}$ & $0.446_{0.11}$ & $0.480_{0.07}$ & $0.482_{0.06}$ & $0.497_{0.07}$ & $0.504_{0.06}$  \\
    \clmsord & $0.440_{0.10}$ & $0.468_{0.11}$ & $0.439_{0.11}$ & $0.515_{0.10}$ & $0.498_{0.09}$ & $0.531_{0.10}$ & $0.556_{0.09}$ \\
    \clmslc & $0.440_{0.10}$ & $0.448_{0.12}$ & $\mathit{0.480_{0.15}}$ & $0.501_{0.09}$ & $0.530_{0.11}$ & $0.520_{0.14}$ & $0.554_{0.11}$ \\
    \bottomrule \bottomrule
    \end{tabular}
    }
\end{table}

The analysis reveals a clear advantage of ordinal methods, particularly the family of CLM-based models, which stand out in the QWK and AMAE metrics (\Cref{tab:results}). In QWK, most of the best-performing configurations correspond to models based on the CLM framework, highlighting its strong suitability for the oak defoliation detection. Notably, CLM achieves the best overall performance in the three-view ensemble, reaching $0.613_{0.10}$. Moreover, second-best values in QWK predominantly arise from ordinal methods (6 out of 7 multi-view configurations), further confirming their robustness. In AMAE, CLM-based models again dominate, especially CLM, which presents $0.513_{0.11}$ in the three-view ensemble. Second-best values in AMAE also mostly correspond to ordinal methods, reinforcing their consistency across evaluation metrics.

As expected, the nominal model shows particularly strong results in terms of accuracy, which aligns with the fact that accuracy is a nominal metric that does not account for the ordinal nature of the labels. This reinforces the inclusion of accuracy in this study, as it provides a fairer basis for evaluating the behaviour of the nominal model, while the ordinal metrics (QWK and AMAE) better capture the advantages of ordinal approaches such as CLM. Although accuracy is a nominal metric and the nominal model performs competitively across several views, it is noteworthy that the best result is obtained with an ordinal method, CDWCE, in the three-view ensemble ($0.579_{0.08}$). 

\begin{figure}[!ht]
    \centering
    \includegraphics[width=\textwidth]{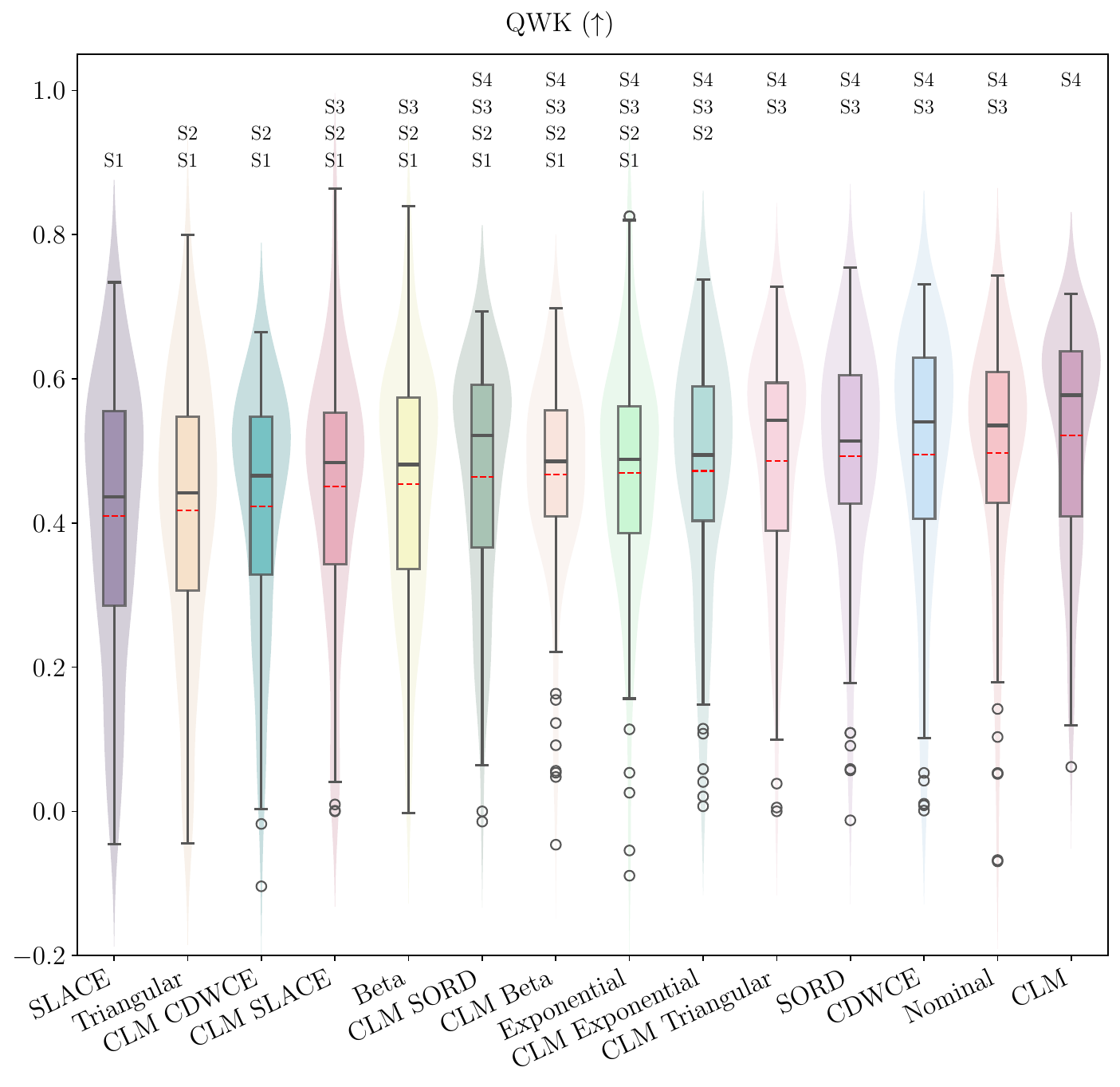}
    \caption{QWK boxplot comparing all methodologies. The solid black line indicates the median, while the dashed red line represents the mean for each boxplot.  Boxplots are ordered according to the Tukey HSD test  from lowest to highest performing methods. Letters above boxplot indicate significantly non-different groups (S1 to S4)}
    \label{fig:QWK_boxplot}
\end{figure}

While CLM emerges as the most consistent and effective methodology across QWK and AMAE, the CDWCE model also achieves competitive performance, obtaining the best accuracy score in the three-view ensemble ($0.579_{0.08}$) and ranking the second best approach in QWK and AMAE. However, the relative strength of CDWCE compared to other ordinal alternatives becomes clearer when considering the global distribution of results across all configurations ( \Cref{fig:QWK_boxplot}).

In the Tukey HSD analysis (\Cref{fig:QWK_boxplot,apx:statistical_analysis}), the CLM method also stands out as the best-performing approach, achieving the highest median ($\sim 0.59$) and mean ($\sim 0.56$) among all methodologies. Its results are tightly concentrated, with minimal dispersion and only one isolated outlier, indicating strong robustness and consistency. In contrast, lower-performing methods such as SLACE and Triangular occupy the initial positions in the figure, corresponding to the lowest Tukey subsets, which confirms their significantly inferior performance. Intermediate methods, including CDWCE, SORD, and several CLM-based variants, form overlapping Tukey subsets, reflecting that their differences are not statistically significant despite exhibiting competitive performance levels. Among them, CDWCE emerges as one of the strongest alternatives, combining relatively high performance with low variability, suggesting stable behaviour across different runs.

\begin{table}[!ht]
    \caption{Mean\textsubscript{std} test results for QWK, AMAE and accuracy for each view across all methodologies. The best result for each metric is shown in bold, and the second-best is shown in italics. Metrics marked with $\uparrow$ are to be maximised and those with $\downarrow$ are to be minimised.}
    \label{tab:mean_results}
    \centering
    \begin{tabular}{lccc}
    \toprule\toprule
    View & QWK $(\uparrow)$ & AMAE $(\downarrow)$ & Accuracy $(\uparrow)$ \\
    \midrule
    Crown & $0.393_{0.043}$ & $0.791_{0.067}$ & $0.385_{0.049}$ \\
    North & $0.397_{0.059}$ & $0.807_{0.078}$ & $0.390_{0.039}$ \\
    South & $0.391_{0.048}$ & $0.817_{0.061}$ & $0.378_{0.035}$ \\
    Crown+North & $\mathit{0.508_{0.032}}$ & $\mathit{0.645_{0.049}}$ & $\mathit{0.457_{0.052}}$ \\
    Crown+South & $0.506_{0.035}$ & $0.646_{0.047}$ & $0.455_{0.046}$ \\
    North+South & $0.507_{0.035}$ & $0.654_{0.040}$ & $0.454_{0.040}$ \\
    Crown+North+South & $\mathbf{0.559_{0.035}}$ & $\mathbf{0.580_{0.047}}$ & $\mathbf{0.493_{0.047}}$ \\
    \bottomrule\bottomrule
    \end{tabular}
\end{table}

\subsection{Performance across view configurations}

The ensemble of three views consistently achieves the best results across the three evaluation metrics (\Cref{tab:mean_results}). This confirms the superiority of considering all available views simultaneously. The second-best results are always obtained with pairwise ensembles, while the use of individual views is consistently less effective. Overall, the pairwise ensembles clearly outperform single-view configurations, highlighting the benefits of combining complementary perspectives.

Among the pairwise ensembles, the Crown+North combination stands out as the most competitive, yielding the second-best overall results across all three metrics. Regarding individual views, no single perspective consistently dominates across metrics. North achieves the best performance in QWK and accuracy, whereas Crown attains the best AMAE value. In contrast, the South view consistently produces the weakest results across the three evaluation measures. Notably, the best pairwise configuration varies depending on the metric: Crown+South achieves the best QWK and AMAE with CDWCE, while Crown+North yields the best accuracy under the nominal model. For single views, North achieves the highest QWK (CLM-Triangular) and Crown the best AMAE and accuracy (nominal model).

\subsection{Performance across sites}

Overall, the Huelva site consistently achieves the best performance across the three performance metrics (QWK, AMAE, and accuracy) (\Cref{tab:plot_analysis}), indicating a more favourable scenario for defoliation estimation. In contrast, Tejera exhibits the lowest performance, with higher variability, suggesting increased difficulty in modelling this site. Almoraima and Córdoba show intermediate behaviour, with competitive results in specific metrics but without consistently outperforming the others. These findings highlight the influence of site-specific conditions on model performance, even when using the same methodology.

\begin{table}[!ht]
    \caption{Mean\textsubscript{std} test results for QWK, AMAE and accuracy across different plots using the best methodology. The best result for each metric is shown in bold, and the second-best is shown in italics. Metrics marked with $\uparrow$ are to be maximised and those with $\downarrow$ are to be minimised.}
    \label{tab:plot_analysis}
    \centering
    \begin{tabular}{lcccc}
    \toprule
    \toprule
    Plot & QWK ($\uparrow$) & AMAE ($\downarrow$) & Accuracy ($\uparrow$) &  \\
    \midrule
    Almoraima & $\mathit{0.614_{0.135}}$ & $\mathit{0.554_{0.151}}$ & $0.480_{0.113}$ &  \\
    Córdoba & $0.577_{0.105}$ & $0.572_{0.164}$ & $\mathit{0.541_{0.133}}$ &  \\
    Huelva & $\mathbf{0.639_{0.168}}$ & $\mathbf{0.368_{0.173}}$ & $\mathbf{0.609_{0.133}}$ &  \\
    Tejera & $0.421_{0.271}$ & $0.589_{0.366}$ & $0.417_{0.255}$ &  \\
    \midrule
    Plot & Sens$_{0}$ ($\uparrow$) & Sens$_{1}$ ($\uparrow$) & Sens$_{2}$ ($\uparrow$) & Sens$_{3}$ ($\uparrow$) \\
    \midrule
    Almoraima & $0.516_{0.357}$ & $0.462_{0.215}$ & $0.320_{0.174}$ & $\mathit{0.749}_{0.184}$ \\
    Córdoba & $\mathit{0.633_{0.287}}$ & $\mathit{0.498_{0.263}}$ & $\mathbf{0.585_{0.270}}$ & $0.443_{0.388}$ \\
    Huelva & $\mathbf{0.772_{0.317}}$ & $\mathbf{0.618_{0.221}}$ & $\mathit{0.493_{0.279}}$ & $\mathbf{0.849_{0.227}}$ \\
    Tejera & $0.462_{0.461}$ & $0.383_{0.380}$ & $0.442_{0.377}$ & $0.667_{0.471}$ \\
    \midrule
    Plot & MAE$_{0}$ ($\downarrow$) & MAE$_{1}$ ($\downarrow$) & MAE$_{2}$ ($\downarrow$) & MAE$_{3}$ ($\downarrow$) \\
    \midrule
    Almoraima & $0.547_{0.408}$ & $0.648_{0.302}$ & $0.681_{0.275}$ & $\mathit{0.301_{0.267}}$ \\
    Córdoba & $\mathit{0.459_{0.302}}$ & $\mathit{0.507_{0.258}}$ & $\mathit{0.512_{0.376}}$ & $0.820_{0.552}$ \\
    Huelva & $\mathbf{0.217_{0.312}}$ & $\mathbf{0.449_{0.259}}$ & $0.598_{0.379}$ & $\mathbf{0.186_{0.298}}$ \\
    Tejera & $0.626_{0.653}$ & $0.687_{0.450}$ & $\mathbf{0.481_{0.381}}$ & $0.333_{0.471}$ \\
    \bottomrule
    \bottomrule
    \end{tabular}
\end{table}

\section{Discussion}\label{sec:discussion}

The results demonstrate that incorporating the ordinal nature of defoliation, coupled with multi-view information, substantially enhances model performance for tree-level health assessment. Across all experiments, ordinal approaches consistently outperformed nominal classifiers when evaluated with order-sensitive metrics (QWK and AMAE). This highlights the criticality of explicitly modelling the inherent ranking of defoliation classes, rather than treating them as independent categories. Such a finding aligns with the interpretation of crown condition as a gradual physiological decline rather than a sequence of discrete states \cite{hartmannClimateChangeRisks2022}. While traditional DL applications in forestry often rely on nominal Cross-Entropy \cite{maDeepLearningRemote2019}, our results corroborate that considering class relationships provides a more nuanced learning signal, consistent with foundational ordinal learning theory \cite{vargas2020cumulative,gutierrez2015ordinal}.

\subsection{Comparison between ordinal and nominal methods}

Among the evaluated methods, CLM-based models emerge as the most robust and consistent approach, achieving the best performance in QWK and AMAE while maintaining low variability across runs. This behaviour is likely explained by their formulation based on an underlying latent continuous variable partitioned by ordered thresholds \cite{agresti2010analysis,vargas2020cumulative}, which naturally encodes the monotonic progression of defoliation levels. By modelling cumulative probabilities rather than treating each class independently, CLM-based models are inherently less likely to producing predictions that violate the ordinal structure of the problem. In contrast, nominal approaches treat class labels as unordered categories and therefore tend to produce less stable predictions when applied to ordinal tasks \cite{gutierrez2015ordinal}. Although alternative ordinal losses such as CDWCE also show competitive performance, particularly in accuracy, their penalisation mechanism operates at the loss level without constraining the output layer, which may limit their ability to fully exploit the ordinal structure during inference. Overall, the results confirm CLM as the most reliable framework for this task. The comparison between other evaluation metrics further supports the relevance of ordinal-aware approaches. Although the nominal model shows competitive performance in accuracy, this metric does not account for the ordered structure of the response variable and may therefore provide a partial assessment of model performance \cite{gutierrez2015ordinal,ferriExperimentalComparisonPerformance2009} . In contrast, QWK and AMAE better reflect the magnitude of prediction errors along the defoliation gradient \cite{baccianella2009evaluation}, providing a more appropriate evaluation framework for this type of problem. 

\subsection{Using multiple visual perspectives}

A consistent performance gain was also observed through the integration of multiple visual perspectives. The three-view ensemble (crown, north, and south) systematically outperformed both single-view and pairwise configurations. This confirms that a multi-perspective approach provides a more holistic representation of the canopy, effectively capturing the structural complexity that single-view observations often miss due to self-shading or foliage clumping \cite{beloiuIndividualTreeCrownDetection2023,kalinDefoliationEstimationForest2019}. This result is particularly relevant when compared with previous image-based approaches for defoliation estimation, which predominantly rely on single-view imagery. For example, \cite{kalinDefoliationEstimationForest2019} demonstrated that convolutional neural networks applied to ground-level images can approximate expert assessments, but their framework is inherently constrained by the perspective from which images are acquired. Similarly, other recent studies using DL for vegetation analysis have shown strong performance at the image level but do not explicitly address the limitations imposed by viewpoint dependency \cite{dacunhaMappingLULCTypes2020,weinsteinCrosssiteLearningDeep2020}. In contrast, our results indicate that explicitly integrating complementary views can significantly reduce this limitation, leading to more robust and stable predictions. While previous studies have noted the spatial heterogeneity of tree crowns \cite{canto-sansoresImportanceSpatialScale2024,lauschUnderstandingForestHealth2017} , the explicit use of multi-view learning strategies remains limited in forest health applications \cite{maDeepLearningRemote2019,weinsteinCrosssiteLearningDeep2020}. This contrasts with other domains of computer vision, where multi-view approaches have been shown to improve the reconstruction of complex three-dimensional structures and reduce uncertainty associated with partial observations \cite{qiVolumetricMultiviewCNNs2016,suMultiviewConvolutionalNeural2015}. Our framework translates these advances into the context of forest health monitoring, where the three-dimensional architecture of tree crowns plays a critical role in determining observable defoliation patterns. Furthermore, the proposed approach mitigates viewpoint-dependent bias and improves robustness against occlusions—a factor of paramount importance in open and structurally complex systems such as Mediterranean dehesas, where asymmetric crown development, crown plasticity, and directional stress responses are prevalent \cite{duchemin2018tree}. These characteristics often lead to uneven foliage distribution, making single-view assessments particularly sensitive to observation angle. Thus, the observed improvements are not merely algorithmic but reflect a better alignment between data acquisition and the ecological reality of the system. Apart from La Tejera, where the low values of QWK and accuracy would be related with the low number of individuals, the best test results correspond to the more homogeneous site. Tree crown assessment is highly influenced by site conditions \cite{eichhornPartIVVisual2020} , so observations are influenced by the overall condition of the area. Consequently, more homogeneous areas will yield more consistent observations, which will also be reflected in the model's accuracy.
It is worth to mention the case of La Almoraima, in which intermediate classes presented a significant low sensitivity compared with extreme classes. This result can be related with the differences between holm and cork oak, regarding leave colour and crown structure. The influence of the woody parts of the crown and the similarity of colour with the foliage in cork oak can lead to more difficulties in the separation in cases of intermediate defoliation.

\section{Conclusions}\label{sec:conclusion}

Overall, these findings demonstrate that combining ordinal classification with a novel multi-view ensemble strategy provides a robust and consistent framework for defoliation estimation. Beyond methodological improvements, this approach has direct implications for forest health monitoring and management, particularly in structurally heterogeneous systems such as Mediterranean dehesas \cite{sanchez-cuestaEnvironmentalDriversInfluencing2021}. By leveraging ground-level imagery and DL, it enables scalable, repeatable, and objective assessments at the individual tree level, reducing observer bias and improving consistency across monitoring campaigns \cite{kalinDefoliationEstimationForest2019, torresRoleRemoteSensing2021}. In this sense, the proposed framework aligns well with existing large-scale monitoring programs, such as ICP Forests, where defoliation is a key indicator but still relies heavily on visual and potentially subjective assessments \cite{eichhornPartIVVisual2020}. From an applied perspective, the proposed framework can support early detection of decline processes by capturing subtle transitions along the defoliation gradient, which are often overlooked in traditional categorical assessments. This is particularly relevant under current global change scenarios, where drought-induced stress and biotic agents interact to drive tree mortality \cite{allenGlobalOverviewDrought2010,hartmannClimateChangeRisks2022}. Furthermore, the integration of multi-view information may enhance the reliability of operational monitoring protocols, providing more robust indicators for decision-making in adaptive forest management. As such, this approach represents a promising step towards bridging the gap between advanced machine learning techniques and practical forest health assessment.

\section*{Acknowledgments}
The present study is supported by the ``Agencia Estatal de Investigación (España)'' (grant ref.: PID2023 - 150663NB - C22 / AEI / 10.13039 / 501100011033), by the EU Commission, AgriFoodTEF (grant ref.: DIGITAL-2022-CLOUD-AI-02, 101100622), by the Secretary of State for Digitalization and AI ENIA International Chair (grant ref.: TSI-100921-2023-3),  LIFE FAGESOS EU project (101074466-LIFE21-CCA-IT-LIFE FAGESOS), DehesAlert project funded by FECYT and Ministerio de Ciencia, Innovación y Universidades (FCT-23-19095) and by the University of Córdoba (grant ref.: PP2F\_L1\_15). F. Bérchez-Moreno has been supported by ``Plan Propio de Investigación Submodalidad 2.2 Contratos predoctorales'' of the University of Córdoba. Víctor Manuel Vargas has been supported by the Ministerio de Ciencia, Innovación y Universidades, the Agencia Estatal de Investigación and the European Social Fund Plus (grant ref.: MICIU / AEI / 10.13039/501100011033, JDC2024-054787-I).  Pablo González Moreno was supported by Grant RYC2021–033138-I, funded by MCIN / AEI / 10.13039 / 501100011033 and European Union (“NextGenerationEU” / PRTR). Ricardo E. Hernández Lambraño was supported by the Grant, JDC2022-050186-I, funded by MCIN / AEI / 10.13039 / 501100011033 and the European Union (“NextGenerationEU” / PRTR). Special thanks to the team who helped us during the field work: R. Sánchez de la Cuesta, A. Molina, A. Ariza, K. Onoszko, M.A. El Chami, J.L. Quero, A. Pérez de Luque, R. Cabrera Puerto, R. Ramírez Luna, P. Sánchez y S. Rodríguez.

\bibliographystyle{elsarticle-num}
\bibliography{bibliography}

\appendix

\section{Components of deep ordinal classification}\label{apx:a-deep-learning}
This appendix provides additional details on the main components used in deep ordinal classification models. In particular, we describe the most common output layer formulations and loss functions employed to incorporate ordinal information into deep learning architectures. These elements complement the overview presented in \Cref{subsubsec:deep_ordinal_classification} and are included here to keep the main manuscript concise.

\subsection{Output Layers}\label{apx:output_layers}
In conventional multi-class classification, the most widely adopted output layer is the softmax function. Given a vector of $J$ real-valued scores produced by the final layer of the network, softmax transforms these scores into a probability distribution over $J$ classes such that the probabilities sum to one. In CNNs, this formulation yields a categorical distribution over the predefined labels, implicitly assuming that the classes are mutually exclusive and unordered.

Although softmax is appropriate for nominal classification tasks, it does not exploit the ordinal relationships between categories. In particular, it does not encode any awareness of proximity between neighbouring classes, assigning equal importance to all categories independently of their position on the ordinal scale. Consequently, errors between adjacent classes are treated equivalently to errors between distant classes.

To address this limitation, ordinal classification is often modelled under the assumption of an underlying latent continuous variable that determines the observed ordered outcomes \cite{agresti2010analysis}. This latent variable can be interpreted as a one-dimensional projection of the input features, obtained through linear or nonlinear transformations. Methods based on this assumption are commonly referred to as threshold models. In these models, the real line is partitioned into $J$ ordered intervals using a strictly increasing sequence of $J-1$ thresholds, and class membership is determined by the interval in which the latent projection lies.

\begin{figure}[!ht]
    \centering
    \includegraphics[width=\textwidth]{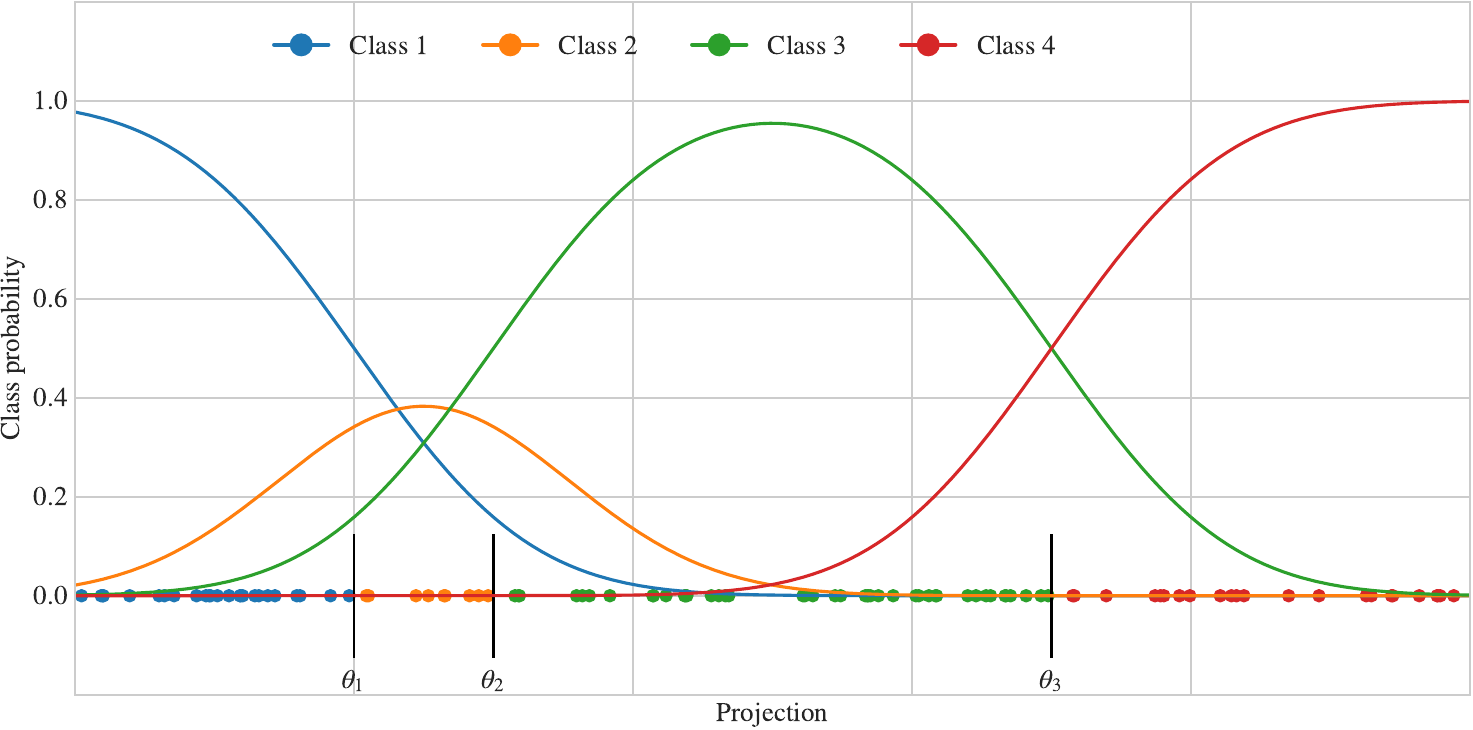}
    \caption{Probability assigned by the CLM to each pattern in a four-class problem, based on its projection. Points on the x-axis represent the projection of each instance on the real line, while the curves indicate the probability assigned to each class. The probability of a pattern belonging to a given class can be obtained by sampling $y$ from the class distribution at the projected $x$ value.}
    \label{fig:clm}
\end{figure}

In neural network implementations, this latent projection corresponds to the scalar output of the final layer, typically represented by a single neuron. The predicted class is obtained by comparing this scalar value against the learned thresholds. Among threshold-based approaches, CLM is one of the most widely used formulations (see \Cref{fig:clm}). The CLM defines a link function that transforms the latent projection into cumulative class probabilities, from which individual class probabilities are derived. The logistic (logit) link is commonly employed, although alternative link functions such as the probit or complementary log-log (cloglog) may also be adopted \cite{vargas2020cumulative}. By imposing ordered thresholds on the latent space, the CLM explicitly incorporates the ordinal structure of the problem into the network’s output layer.

\subsection{Loss functions}\label{apx:loss_functions}
Loss functions define the optimisation objective that guides the training of deep neural networks. In classification tasks, they quantify the discrepancy between the predicted probability distribution and the ground-truth label, defining the objective function that guides the optimisation of the model parameters. The choice of loss function is therefore a central component of deep ordinal classification, as it determines how prediction errors are penalised during training.

In nominal classification tasks with unordered classes, the most widely used loss function is the Categorical Cross-Entropy (CCE). This function measures the dissimilarity between the predicted probability distribution and the target label distribution. Given an input $\mathbf{x}_i$, the loss associated with predicting class $\class{k}$ is expressed as:

\begin{equation}
        \mathscr{L}\left(x, k\right) = \sum^{J}_{j=1} h \left(j,k\right)\left(-\log p\left(y = \class{j} | \mathbf{x}_i\right)\right),
\end{equation}
where $J$ denotes the total number of classes and $h(j, k)$ represents the target distribution over labels. For hard labels encoded using a one-hot representation, this function reduces to the indicator:

\begin{equation}
    h(j, k) = \mathds{1}\{j = k\},
\end{equation}
which equals $1$ when the predicted class index $j$ matches the true class index $k$, and $0$ otherwise.

Here, $p(y = \class{j} \mid \mathbf{x}_i)$ is the predicted probability assigned to class $\class{j}$ for the input $\mathbf{x}_i$, typically obtained from a softmax layer. This formulation assumes no inherent order within the classes and penalises all misclassifications equally, regardless of their relative position on any ordinal scale.

However, in OC problems, such as defoliation severity estimation, this assumption is inappropriate. Predicting a neighbouring category should not be penalised as severely as predicting a distant one. To better exploit the ordinal structure of the labels, several alternative loss functions incorporate distance-aware mechanisms that penalise errors progressively according to the ordinal distance between the predicted and true classes. These approaches belong to the family of distance-based ordinal losses.

Among distance-based ordinal losses, several formulations incorporate class proximity directly into the optimisation objective. The Class Distance-Weighted Cross-Entropy (CDWCE) loss \cite{polat2022class} extends the standard cross-entropy by introducing a distance-dependent weighting factor, assigning progressively larger penalties as the ordinal gap between predicted and true classes increases. The Soft Ordinal Regression (SORD) loss \cite{diaz2019soft} replaces the one-hot target encoding with a smooth probabilistic label distribution derived from the distances between ranks, introducing smooth target distributions that reflect ordinal proximity, without requiring architectural modifications. Finally, the SLACE loss \cite{nachmani2025slace} builds upon cumulative probability formulations and incorporates proximity-aware weighting that accounts for both ordinal distance and class distribution characteristics, promoting ordinal consistency while remaining robust to class imbalance.

Collectively, these distance-based loss formulations enhance the model’s ability to exploit the ordinal structure of the target labels by embedding structured inter-class relationships directly into the optimisation objective. This leads to more coherent learning dynamics and smoother decision boundaries, as analysed in the comparative experiments presented in \Cref{sec:results}.

\subsection{Soft Labelling}\label{apx:soft_labelling}
Beyond distance-based ordinal losses, another family of ordinal-aware alternatives is given by SL strategies, which aim to relax the strictness of one-hot encoding by introducing structured probabilistic targets. Instead of assigning full probability mass to a single class, SL represents each instance by a probability distribution over the ordinal categories. This approach allows a certain degree of tolerance to near-miss predictions and better reflects the inherent uncertainty present in many ordinal problems.

In conventional multiclass classification, prior to introducing ordinal-specific SL schemes, a common strategy to mitigate overconfident predictions consists of incorporating regularisation based on a uniform distribution within the loss function \cite{szegedy2016rethinking}. The resulting loss can be written as:

\begin{equation}\label{eq:3}
    \mathscr{L}\left(x, k\right) = \sum^{J}_{j=1} h' \left(j,k\right)\left(-\log p\left(y = \class{j} | \mathbf{x}_i\right)\right),
\end{equation}
where $h' \left(j,k\right) = (1 - \lambda) \mathds{1}\{j = k\} + \lambda\frac{1}{J}$. Here, $h'\left(j,k\right)$ denotes a convex combination of a one-hot (hard) label and a uniform (soft) distribution, controlled by the hyperparameter $\lambda \in [0,1]$.

For ordinal classification tasks, this regularisation scheme can be adapted by replacing the uniform distribution with a unimodal ordinal distribution that reflects class proximity. Unlike the uniform case, where all categories are treated equally, ordinal scenarios require a structure that assigns higher probability to neighbouring classes and lower probability to distant ones. In this context, the weighting term $h'(j,k)$ in \Cref{eq:3} is replaced by:

\begin{equation}
    h'' \left(j,k\right) = (1 - \lambda) \mathds{1}\{j = k\} + \lambda P\left(y = \class{j} | \class{k} \right), \quad \lambda\in[0,1],
\end{equation}
where $\prob{y=\class{j} \mid \class{k}}$ represents the probability that a sample from class $\class{k}$ belongs to class $\class{j}$. This probability can be obtained from the probability density function (p.d.f.) of a continuous distribution, such as the beta or triangular distribution, or from the probability mass function (p.m.f.) of a discrete distribution, such as the exponential distribution.

For discrete distributions, $\prob{y=\class{j} \mid \class{k}}$ is directly given by the p.m.f. In contrast, for continuous distributions, it can be computed as:

\begin{equation}\label{eq:slintegral}
    \prob{y = \class{i} \mid \class{j}} = \int\displaylimits_{(i-1)/J}^{i/J} f_j(u) \, \diff{u},
\end{equation}
where $f_j(u)$ denotes the p.d.f. associated with class $\class{j}$, evaluated at $u$. The integral limits, $(i-1)/J$ and $i/J$, partition the interval $[0,1]$ into $J$ equal segments, each corresponding to one of the $J$ classes, as described in \cite{vargas2024ebano}. This encoding allows the loss to account for the uncertainty associated with ordinal problems by assigning more weight to labels closer to the true class.

Within this framework, several unimodal distributions have been proposed to model $\prob{y=\class{j} \mid \class{k}}$, including triangular \cite{vargas2023soft}, beta \cite{vargas2022unimodal}, and exponential \cite{vargas2023exponential} formulations. The triangular strategy defines class probabilities using a triangular distribution centred on the annotated label, producing a linear decay towards neighbouring categories and adapting its shape at the ordinal boundaries. The beta-based approach introduces a flexible two-parameter distribution that controls the concentration and spread of probability mass, allowing different confidence levels to be represented across samples. Similarly, the exponential formulation models class probabilities through a distance-dependent decay governed by a tunable $L_p$-norm parameter, enabling control over how sharply probabilities decrease as ordinal distance increases.

\begin{figure}[!ht]
    \centering
    \begin{subfigure}{0.45\textwidth}
        \centering
        \includegraphics[width=\linewidth]{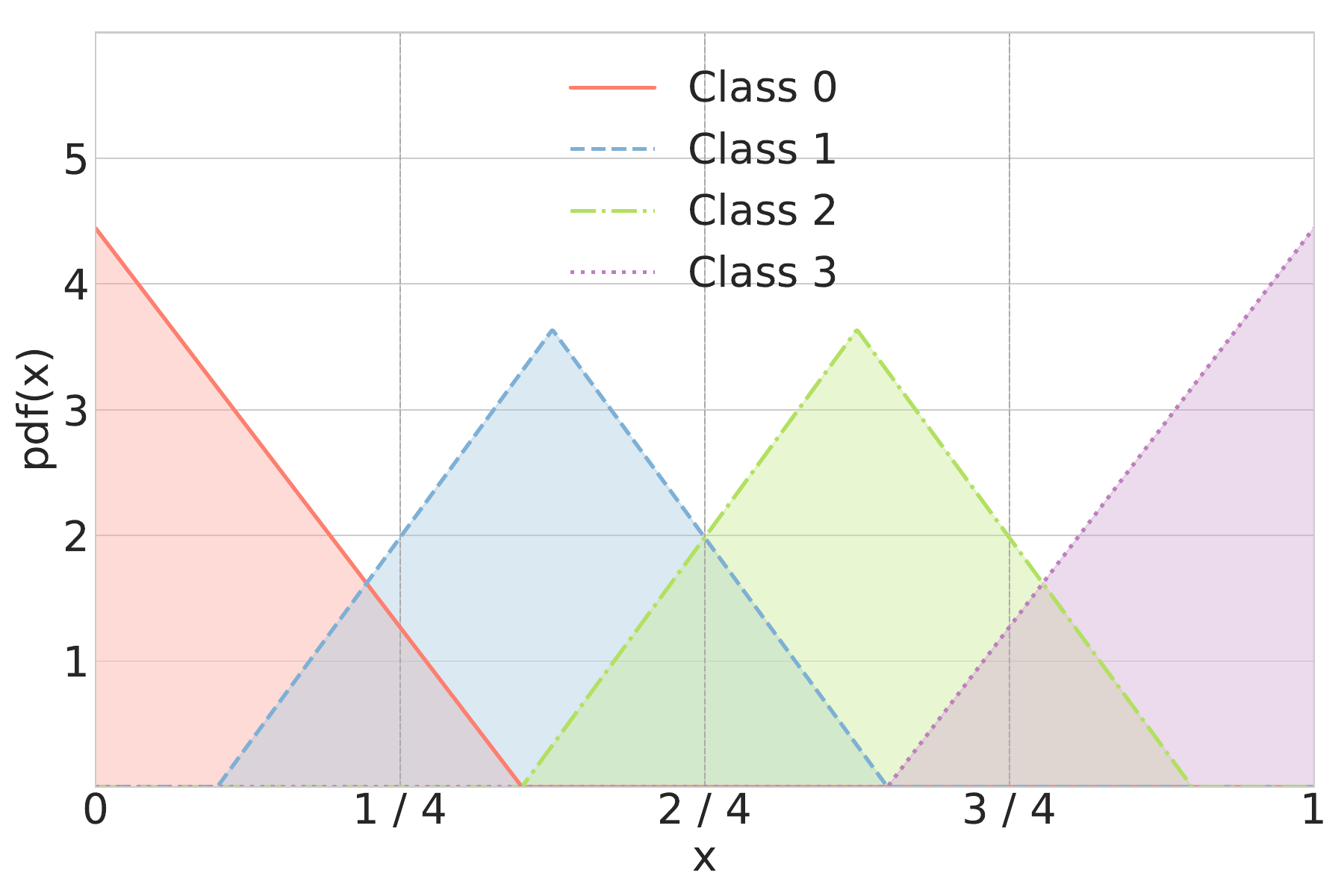}
        \caption{Triangular}
        \label{fig:triangular_distribution}
    \end{subfigure}
    \hfill
    \begin{subfigure}{0.45\textwidth}
        \centering
        \includegraphics[width=\linewidth]{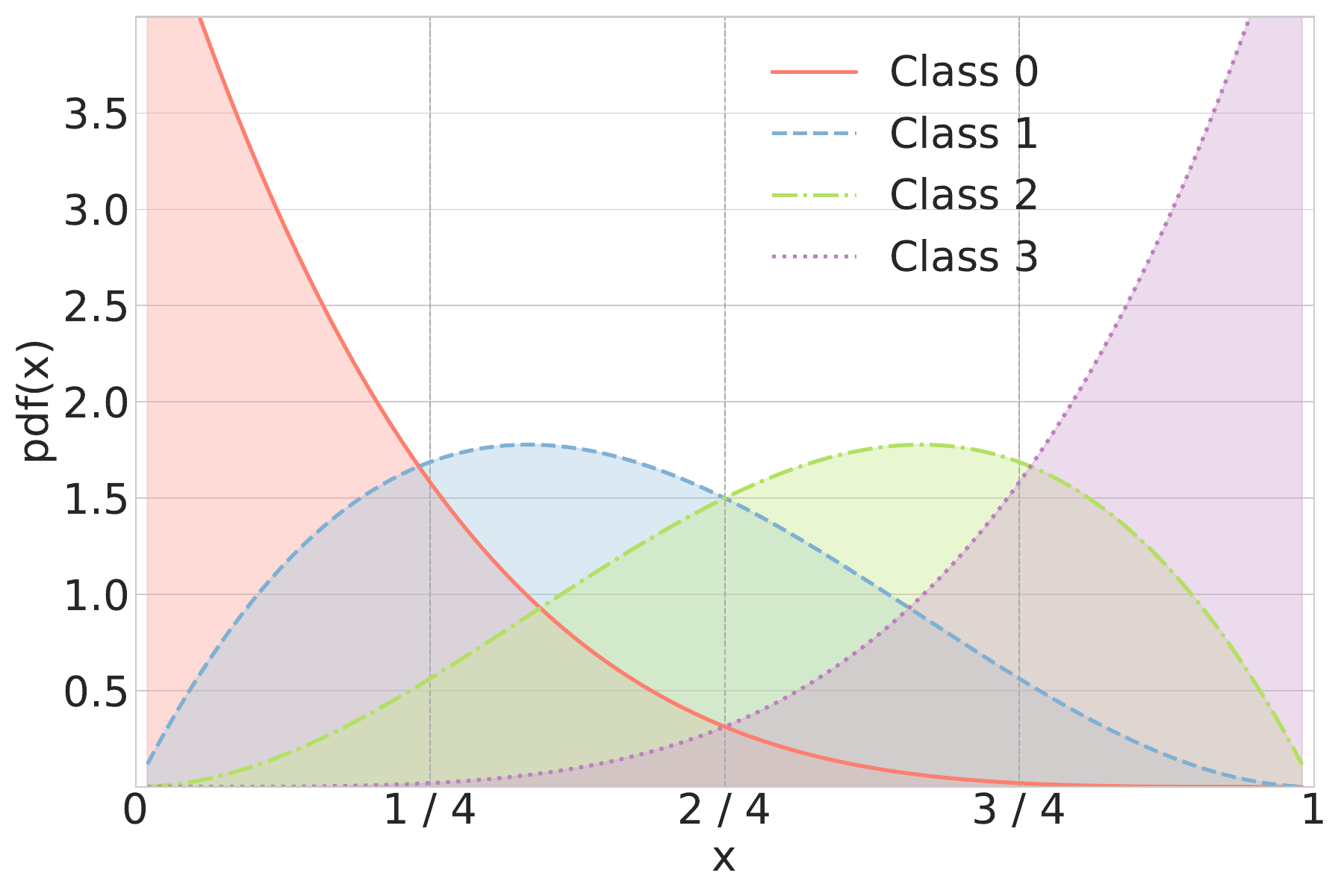}
        \caption{Beta}
        \label{fig:beta_distribution}
    \end{subfigure}

    \vspace{1em}
    \begin{subfigure}{0.6\textwidth}
        \centering
        \includegraphics[width=\linewidth]{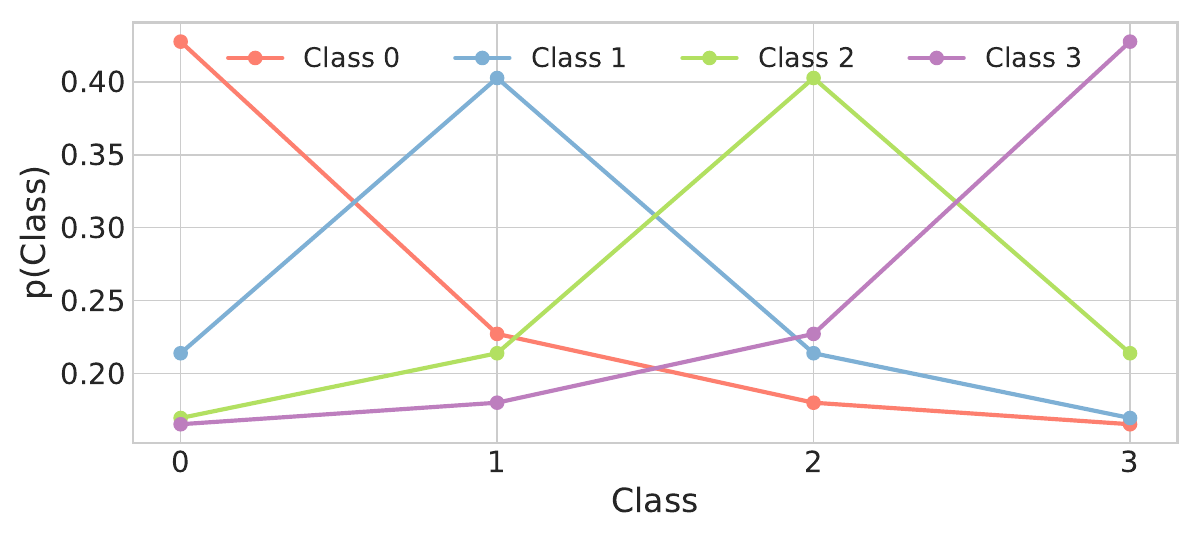}
        \caption{Exponential}
        \label{fig:exponential_distribution}
    \end{subfigure}
    
    \caption{Representation of unimodal distributions}
    \label{fig:unimodal_distributions}
\end{figure}

In this work, we address an OC problem involving oak defoliation detection, a task characterised by subtle visual differences between adjacent severity levels and by inter-observer variability in the annotation process. Misclassifications are therefore more likely to occur between neighbouring categories than between extreme ones. SL strategies provide a natural mechanism to model this uncertainty. For example, in a four-class ordinal problem, a sample belonging to the third class would be encoded under one-hot labelling as $(0,0,1,0)$, whereas a SL strategy may represent it as $(0,0.2,0.7,0.1)$, distributing probability mass across adjacent levels according to their ordinal proximity. The corresponding unimodal distributions are illustrated in \Cref{fig:unimodal_distributions}.

\section{Hyperparameter tuning process}\label{apx:crossvalidation}
This appendix provides a detailed account of the hyperparameter tuning process employed in the experiments. The procedure involved cross-validation guided by the AMAE metric \cite{baccianella2009evaluation}, as introduced previously in this work. A formal definition of AMAE and the specific search strategy used for each methodology are presented below.

\begin{equation}
    \text{AMAE} = \frac{1}{J} \sum_{j=1}^J \frac{1}{N_j} \sum_{i=1}^{N_j} \left| \order{y_i} - \order{\hat{y}_i} \right| \quad \in [0, J - 1],
\end{equation}
where $N_j$ corresponds to the number of samples belonging to class $\class{j}$, and $\hat{y}_i$ denotes the predicted ordinal label for the $i$-th instance. Lower AMAE values indicate better performance.

For each seed, a randomised search was performed over 15 iterations to identify the optimal combination of hyperparameters. When the total number of possible hyperparameter combinations was fewer than 15, an exhaustive grid search was conducted instead. The ranges of values considered for each hyperparameter during the tuning process are detailed in \Cref{tab:parameters}.

\begin{table}[!ht]
    \scriptsize
    \centering
    \caption{Considered range of values for tuning the hyperparameters of each methodology.}
    \label{tab:parameters}
    \begin{tabular}{llc}
        \toprule \toprule
        Classifier(s) & Hyperparameter(s) & Range of values \\
        \midrule
        \nom & Learning rate & $\{10^{-4},10^{-3},10^{-2}\}$ \\ [0.15cm]

        \multirow{3}{*}{\triang} & Learning rate & $\{10^{-4},10^{-3},10^{-2}\}$ \\
        & Adjacent class probability & $\{0.01, 0.05, 0.10\}$ \\
        & Smoothing factor $\eta$ & $\{0.8,1.0\}$ \\ [0.15cm]

        \multirow{2}{*}{\bet} & Learning rate & $\{10^{-4},10^{-3},10^{-2}\}$ \\
        & Smoothing factor $\eta$ & $\{0.8, 1.0\}$ \\ [0.15cm]

        \multirow{3}{*}{\expo} & Learning rate & $\{10^{-4},10^{-3},10^{-2}\}$ \\
        & Smoothing factor $\eta$ & $\{0.8, 1.0\}$ \\
        & Exponent & $\{1.0, 1.5, 2.0\}$ \\ [0.15cm]

        \multirow{2}{*}{\cdw} & Learning rate & $\{10^{-4},10^{-3},10^{-2}\}$ \\
        & Exponent & $\{0.25, 0.50, 0.75, 1.00\}$ \\ [0.15cm]

        \multirow{2}{*}{\slc} & Learning rate & $\{10^{-4},10^{-3},10^{-2}\}$ \\
        & Smoothing factor & $\{1, 0.3, 0.5, 0.8, 2, 3, 4, 7, 10, 15, 20, 25\}$ \\ [0.15cm]

        \multirow{4}{*}{\sord} & Learning rate & $\{10^{-4},10^{-3},10^{-2}\}$ \\
        & Smoothing factor & $\{0.3, 0.5, 0.8, 1, 2, 3, 4, 7, 10, 15, 20, 25\}$ \\
        & \multirow{2}{*}{Proximity matrix transformation} & \{max, norm max, norm log, \\
        &  & log, norm division, division\} \\ [0.15cm]

        \multirow{2}{*}{\clm} & Learning rate & $\{10^{-4},10^{-3},10^{-2}\}$ \\
        & Minimum distance & $\{0.0, 0.5, 1.0\}$ \\ [0.15cm]

        \multirow{3}{*}{\clmbeta} & Learning rate & $\{10^{-4},10^{-3},10^{-2}\}$ \\
        & Smoothing factor $\eta$ & $\{0.8, 1.0\}$ \\
        & Minimum distance & $\{0.0, 0.5, 1.0\}$ \\ [0.15cm]

        \multirow{4}{*}{\clmtriang} & Learning rate & $\{10^{-4},10^{-3},10^{-2}\}$ \\
        & Smoothing factor $\eta$ & $\{0.8, 1.0\}$ \\
        & Adjacent class probability & $\{0.01, 0.05, 0.10\}$ \\
        & Minimum distance & $\{0.0, 0.5, 1.0\}$ \\ [0.15cm]

        \multirow{4}{*}{\clmexp} & Learning rate & $\{10^{-4},10^{-3},10^{-2}\}$ \\
        & Smoothing factor $\eta$ & $\{0.8, 1.0\}$ \\
        & Minimum distance & $\{0.0, 0.5, 1.0\}$ \\
        & Exponent & $\{1.0, 1.5, 2.0\}$ \\ [0.15cm]

        \multirow{3}{*}{\clmcdwce} & Learning rate & $\{10^{-4},10^{-3},10^{-2}\}$ \\
        & Exponent & $\{0.25, 0.50, 0.75, 1.00\}$ \\
        & Minimum distance & $\{0.0, 0.5, 1.0\}$ \\ [0.15cm]

        \multirow{3}{*}{\clmslc} & Learning rate & $\{10^{-4},10^{-3},10^{-2}\}$ \\
        & Smoothing factor & $\{1, 0.3, 0.5, 0.8, 2, 3, 4, 7, 10, 15, 20, 25\}$ \\
        & Minimum distance & $\{0.0, 0.5, 1.0\}$ \\ [0.15cm]

        \multirow{5}{*}{\clmsord} & Learning rate & $\{10^{-4},10^{-3},10^{-2}\}$ \\
        & Smoothing factor & $\{0.3, 0.5, 0.8, 1, 2, 3, 4, 7, 10, 15, 20, 25\}$ \\
        & \multirow{2}{*}{Proximity matrix transformation} & \{max, norm max, norm log, \\
        &  & log, norm division, division\} \\
        & Minimum distance & $\{0.0, 0.5, 1.0\}$ \\ [0.15cm]
        
        \bottomrule \bottomrule
    \end{tabular}
\end{table}

\section{Evaluation metrics}\label{apx:metrics}
In this appendix, we provide a detailed description of the evaluation metrics employed throughout this study. These metrics were selected to assess both the ordinal and nominal aspects of the oak defoliation detection. For ordinal evaluation, metrics such as QWK and AMAE are considered, while standard nominal metrics, including the accuracy, provide a complementary perspective on overall predictive performance. The formal definitions and computational details of each metric are presented below.

\begin{itemize}
    \item Quadratic Weighted Kappa (QWK) \cite{AYLLONGAVILAN2026112273} is a metric specifically designed for ordinal classification tasks, as it accounts for the magnitude of disagreement between predicted and true classes. It penalises more heavily those predictions that are further from the actual label. The QWK is defined as:
        \begin{equation}
            \text{QWK} = 1 - \frac{\sum\limits_{i=1}^{J} \sum\limits_{j=1}^{J} \omega_{ij} O_{ij}}{\sum\limits_{i=1}^{J} \sum\limits_{j=1}^{J} \omega_{ij} E_{ij}},
        \end{equation}    
    where $J$ represents the total number of ordinal categories. The elements $\omega_{ij}$ of the penalisation matrix represent the cost associated with misclassifying class $\class{i}$ as class $\class{j}$. For ordinal tasks, it is common to define these elements as:
        \begin{equation}
            \omega_{ij} = \frac{|i - j|^n}{(J - 1)^n},
        \end{equation}      
    where $n$ determines the exponent of the penalty. A value of $n=1$ corresponds to a linear penalty, while $n=2$ results in a quadratic penalty, which is typically preferred in ordinal settings to emphasise more severe misclassifications.

    The observed confusion matrix is denoted by $\mathbf{O}$, where $O_{ij}$ represents the number of samples belonging to class $i$ that have been predicted as class $j$. The expected confusion matrix under the assumption of random predictions, denoted $E_{ij}$, is computed as:
        \begin{equation}
            E_{ij} = \frac{O_{i \bullet} O_{\bullet j}}{J},
        \end{equation}  
    where $O_{i \bullet}$ is the sum of the $i$-th row and $O_{\bullet j}$ is the sum of the $j$-th column in the observed matrix $\mathbf{O}$.

    \item Accuracy \cite{rosati2022novel}, quantifies the proportion of samples that are assigned to their correct class. It can be formally defined as:
        \begin{equation}
            \text{Accuracy} = \frac{1}{N} \sum_{j=1}^{J} O_{jj},
        \end{equation}      
    where $O_{jj}$ is the $j$-th element of the main diagonal of the confusion matrix, $J$ is the number of classes, and $N$ is the total number of patterns. 
\end{itemize}

\section{Statistical Analysis}\label{apx:statistical_analysis}
In this appendix, we provide a detailed account of the statistical analyses conducted to assess the performance differences across methods and dataset configurations. Specifically, we report the results of two-way ANOVA and subsequent Tukey HSD posthoc tests for the evaluation metrics considered in this work: QWK, AMAE, and Accuracy. These analyses allow for a rigorous comparison of methods, highlighting statistically significant differences both across models and dataset subsets.

\subsection{Statistical Analysis for QWK}
To assess whether the differences in the QWK metric were statistically significant across the considered factors, we conducted a two-way ANOVA (ANOVA II) with Method and View as the independent variables. The results are reported in \Cref{tab:anova_qwk}. The analysis revealed a significant main effect of both Method ($p<0.001$) and View ($p<0.001$) on the QWK metric. These findings confirm that the choice of method as well as the selection of dataset (i.e. the specific view or combination of views) individually lead to statistically significant differences in performance. In contrast, no significant interaction effect was found between Method and View ($p=0.979$), suggesting that their influences on QWK operate independently rather than in combination.

\begin{table}[!ht]
    \centering
    \caption{Results of the ANOVA II test for the QWK metric. SS: Sum of Squares, DF: Degrees of Freedom, F: F-statistic.}
    \label{tab:anova_qwk}

    \begin{tabular}{lcccc}
    \toprule\toprule
     & SS & DF & F & $p$-value \\
    \midrule
    Method & 1.920 & 13.000 & 6.205 & $<0.001$\\
    View & 8.226 & 6.000 & 57.601 & $<0.001$ \\
    Method * View & 1.298 & 78.000 & 0.699 & 0.979 \\
    Residual & 44.317 & 1862.000 &  &  \\
    \bottomrule\bottomrule
    \end{tabular}
\end{table}

Following the ANOVA results, we applied Tukey's HSD posthoc tests to further explore pairwise differences within each factor (see \Cref{tab:tukey_qwk_estimator,tab:tukey_qwk_dataset}). Regarding the Method factor, the grouping analysis revealed that the CLM model achieved the highest QWK mean value ($0.521$), forming a distinct upper group compared to lower-performing methods such as SLACE and Triangular. Several intermediate models (e.g., CLM Beta, Exponential, CLM Exponential) showed competitive performance, their differences were not statistically significant with respect to CLM or among themselves, suggesting a similar overall behaviour.

In contrast, for the View factor, the results show clear and statistically significant groupings based on performance levels: single datasets (South, North, Crown) formed the lower group with QWK means around $0.391$–$0.397$, while pairwise combinations (Crown+South, North+South, Crown+North) constituted an intermediate group. The three-view ensemble (Crown+North+South) formed the best group, achieving the highest mean QWK of $0.559$.

\begin{table}[!ht]
\centering
\caption{Tukey test results for the QWK metric in terms of the two factors considered.}

    \begin{subtable}{0.42\textwidth}
        \centering
        \caption{Method}
        \tiny

    \begin{tabular}{lcccc}
    \toprule\toprule
     & S1 & S2 & S3 & S4\\
    Group &  &  &  & \\
    \midrule
    \slc & 0.410 &  &  & \\
    \triang & 0.418 & 0.418 &  & \\
    \clmcdwce & 0.423 & 0.423 &  & \\
    \clmslc & 0.451 & 0.451 & 0.451 & \\
    \bet & 0.454 & 0.454 & 0.454 & \\
    \clmsord & 0.464 & 0.464 & 0.464 & 0.464\\
    \clmbeta & 0.468 & 0.468 & 0.468 & 0.468\\
    \expo & 0.469 & 0.469 & 0.469 & 0.469\\
    \clmexp &  & 0.472 & 0.472 & 0.472\\
    \clmtriang &  &  & 0.486 & 0.486\\
    \sord &  &  & 0.493 & 0.493\\
    \cdw &  &  & 0.495 & 0.495\\
    \nom &  &  & 0.497 & 0.497\\
    \clm &  &  &  & 0.521\\ 
    \bottomrule\bottomrule
    \end{tabular}
    \label{tab:tukey_qwk_estimator}
    \end{subtable}
    \hspace{0.7cm}
    \begin{subtable}{0.5\textwidth}
        \centering
        \caption{Dataset}

        \tiny
        \begin{tabular}{lccc}
        \toprule\toprule
         & S1 & S2 & S3 \\
        Group &  &  &  \\
        \midrule
        South & 0.391 &  &  \\
        Crown & 0.393 &  &  \\
        North & 0.397 &  &  \\
        Crown+South &  & 0.506 &  \\
        North+South &  & 0.507 &  \\
        Crown+North &  & 0.508 &  \\
        Crown+North+South &  &  & 0.559 \\
        \bottomrule\bottomrule
        \end{tabular}
        \label{tab:tukey_qwk_dataset}
    \end{subtable}\\[0.3cm]
\end{table}

\subsection{Statistical Analysis for AMAE}

To assess whether the differences in the AMAE metric were statistically significant across the considered factors, we conducted a two-way ANOVA (ANOVA II) with Method and View as the independent variables. The results are reported in \Cref{tab:anova_amae}. The analysis revealed a significant main effect of both Method ($p<0.001$) and View ($p<0.001$) on the AMAE metric. These findings confirm that the choice of method as well as the selection of dataset individually lead to statistically significant differences in performance. In contrast, no significant interaction effect was found between Method and View ($p=0.990$), suggesting that their influences on AMAE operate independently rather than in combination.

\begin{table}[!ht]
    \centering
    \caption{Results of the ANOVA II test for the AMAE metric. SS: Sum of Squares, DF: Degrees of Freedom, F: F-statistic.}
    \label{tab:anova_amae}

    \begin{tabular}{lcccc}
    \toprule\toprule
     & SS & DF & F & $p$-value \\
    \midrule
    Method & 3.573 & 13.000 & 6.029 & $<0.001$\\
    View & 15.614 & 6.000 & 57.095 & $<0.001$ \\
    Method * View & 2.349 & 78.000 & 0.661 & 0.990 \\
    Residual & 84.868 & 1862.000 &  &  \\
    \bottomrule\bottomrule
    \end{tabular}
\end{table}

Following the ANOVA results, we applied Tukey's HSD posthoc tests to further explore pairwise differences within each factor (see \Cref{tab:tukey_amae_estimator,tab:tukey_amae_dataset}). Regarding the Method factor, the grouping analysis revealed that the nominal approach achieved the lowest AMAE value ($0.632$), forming the best-performing group. The CLM model followed closely with an AMAE of $0.636$, while methods such as SORD and CDWCE also showed competitive performance. In contrast, approaches such as SLACE and Triangular obtained the highest AMAE values, indicating poorer performance under this metric.

For the Dataset factor, the results show clear group separations. Single views (South, North, Crown) produced the highest AMAE values, indicating worse performance. Pairwise combinations (Crown+South, North+South, Crown+North) formed an intermediate group with improved results. The three-view combination (Crown+North+South) achieved the lowest AMAE ($0.580$), confirming that combining all available views provides the best predictive performance.

\begin{table}[!ht]
\centering
\caption{Tukey test results for the AMAE metric in terms of the two factors considered.}

    \begin{subtable}{0.42\textwidth}
        \centering
        \caption{Method}
        \tiny

    \begin{tabular}{lcccc}
    \toprule\toprule
     & S1 & S2 & S3 & S4\\
    Group &  &  &  & \\
    \midrule
    \clmcdwce & 0.781 &  &  & \\
    \triang & 0.754 &  &  & \\
    \slc & 0.750 & &  & \\
    \clmslc & 0.734 & 0.734 &  & \\
    \bet & 0.721 & 0.721 & 0.721 & \\
    \clmbeta & 0.720 & 0.720 & 0.720 & \\
    \clmexp & 0.712 & 0.712 & 0.712 & 0.712\\
    \clmsord & 0.709 & 0.709 & 0.709 & 0.709\\
    \clmtriang & 0.704 & 0.704 & 0.704 & 0.704\\
    \expo & 0.703 & 0.703 & 0.703 & 0.703\\
    \cdw &  & 0.662 & 0.662 & 0.662\\
    \sord &  & 0.660 & 0.660 & 0.660\\
    \clm &  &  & 0.636 & 0.636\\
    \nom &  &  &  & 0.632\\ 
    \bottomrule\bottomrule
    \end{tabular}
    \label{tab:tukey_amae_estimator}
    \end{subtable}
    \hspace{0.7cm}
    \begin{subtable}{0.5\textwidth}
        \centering
        \caption{Dataset}

        \tiny
        \begin{tabular}{lccc}
        \toprule\toprule
         & S1 & S2 & S3 \\
        Group &  &  &  \\
        \midrule
        South & 0.817 &  &  \\
        North & 0.807 &  &  \\
        Crown & 0.791 &  &  \\
        North+South &  & 0.654 &  \\
        Crown+South &  & 0.646 &  \\
        Crown+North &  & 0.645 &  \\
        Crown+North+South &  &  & 0.580 \\
        \bottomrule\bottomrule
        \end{tabular}
        \label{tab:tukey_amae_dataset}
    \end{subtable}\\[0.3cm]
\end{table}

\subsection{Statistical Analysis for Accuracy}

To analyse the statistical significance of the differences observed in the Accuracy metric, we performed a two-way ANOVA (ANOVA II) considering Method and View as independent variables. The results are summarised in \Cref{tab:anova_accuracy}. The analysis revealed significant main effects for both Method ($p<0.001$) and View ($p<0.001$), indicating that both the learning approach and the selected dataset configuration significantly influence classification accuracy. No significant interaction effect between Method and View was observed ($p = 0.999$), suggesting that their effects on accuracy are independent.

\begin{table}[!ht]
    \centering
    \caption{Results of the ANOVA II test for the Accuracy metric. SS: Sum of Squares, DF: Degrees of Freedom, F: F-statistic.}
    \label{tab:anova_accuracy}

    \begin{tabular}{lcccc}
    \toprule\toprule
     & SS & DF & F & $p$-value \\
    \midrule
    Method & 3.037 & 13.000 & 18.135 & $<0.001$\\
    View & 3.400 & 6.000 & 43.983 & $<0.001$ \\
    Method * View & 0.574 & 78.000 & 0.571 & 0.999 \\
    Residual & 23.990 & 1862.000 &  &  \\
    \bottomrule\bottomrule
    \end{tabular}
\end{table}

Following the ANOVA results, Tukey’s HSD posthoc tests were conducted to further analyse pairwise differences between methods and datasets (see \Cref{tab:tukey_accuracy_estimator,tab:tukey_accuracy_dataset}). For the Method factor, the results reveal a gradual transition across the Tukey subsets from S1 to S7. The CLM CDWCE variant appears in the lowest-performing subset (S1), presenting the smallest accuracy value ($0.363$). Other CLM-based variants combined with different loss functions, such as CLM SLACE, CLM Beta, and CLM Exponential, also appear in the lower subsets, indicating comparatively poorer performance in terms of accuracy.

Intermediate subsets include models based on exponential, beta, and triangular losses, as well as SLACE and the CLM model, which progressively achieve higher accuracy values. The SORD and CDWCE models occupy the upper subsets of the table, reflecting competitive performance levels.

Finally, the nominal classifier achieves the highest accuracy value ($0.507$) and forms the best-performing subset (S7). However, the overlap between subsets indicates that several models share statistically indistinguishable performance levels according to the Tukey test.

Regarding the Dataset factor, the subsets show a clear improvement in accuracy as more views are combined. Single views (South, Crown, and North) form the lowest-performing subset (S1), while pairwise combinations occupy the intermediate subset (S2). The combination of all views (Crown+North+South) achieves the highest accuracy value ($0.493$) and forms the best-performing subset (S3), suggesting that incorporating information from multiple regions leads to improved classification accuracy.

\begin{table}[!ht]
    \centering
    \caption{Tukey test results for the Accuracy metric in terms of the two factors considered.}
    
    \begin{subtable}{0.8\textwidth}
        \centering
        \caption{Method}
        \tiny
        \begin{tabular}{lccccccc}
        \toprule\toprule
         & S1 & S2 & S3 & S4 & S5 & S6 & S7\\
        Group &  &  &  &  &  &  & \\
        \midrule
        \clmcdwce  & 0.363 &  &  &  &  &  &  \\
        \clmslc    & 0.392 & 0.392 &  &  &  &  &  \\
        \clmbeta   & 0.398 & 0.398 &  &  &  &  &  \\
        \clmexp    & 0.399 & 0.399 &  &  &  &  &  \\
        \clmtriang & 0.403 & 0.403 & 0.403 &  &  &  &  \\
        \clmsord   &  & 0.409 & 0.409 &  &  &  &  \\
        \expo      &  & 0.416 & 0.416 & 0.416 &  &  &  \\
        \bet       &  & 0.435 & 0.435 & 0.435 &  &  &  \\
        \triang    &  & 0.435 & 0.435 & 0.435 & 0.435 &  &  \\
        \slc       &  &  & 0.445 & 0.445 & 0.445 & 0.445 &  \\
        \clm       &  &  &  & 0.460 & 0.460 & 0.460 &  \\
        \sord      &  &  &  & 0.481 & 0.481 & 0.481 &  \\
        \cdw       &  &  &  &  &  & 0.483 & 0.483 \\
        \nom       &  &  &  &  &  &  & 0.507 \\
        \bottomrule\bottomrule
        \end{tabular}
        \label{tab:tukey_accuracy_estimator}
        \end{subtable}
        
        \vspace{0.5cm}
        
        \begin{subtable}{0.6\textwidth}
        \centering
        \caption{Dataset}
        \tiny
        \begin{tabular}{lccc}
        \toprule\toprule
         & S1 & S2 & S3 \\
        Group &  &  &  \\
        \midrule
        South & 0.378 &  &  \\
        Crown & 0.385 &  &  \\
        North & 0.390 &  &  \\
        North+South &  & 0.454 &  \\
        Crown+South &  & 0.455 &  \\
        Crown+North &  & 0.457 &  \\
        Crown+North+South &  &  & 0.493 \\
        \bottomrule\bottomrule
        \end{tabular}
        \label{tab:tukey_accuracy_dataset}
    \end{subtable}
\end{table}

\end{document}